\documentclass[letterpaper]{article} 
\usepackage{aaai2026}  
\usepackage{times}  
\usepackage{helvet}  
\usepackage{courier}  
\usepackage[hyphens]{url}  
\usepackage{graphicx} 
\usepackage{natbib}  
\usepackage{caption} 
\usepackage{algorithm}
\usepackage{algorithmic}
\usepackage{cite}
\usepackage{amsmath,amssymb,amsfonts}
\usepackage{textcomp}
\usepackage{multirow}
\usepackage{booktabs}
\usepackage{pifont}           
\usepackage{array}
\usepackage{tablefootnote}
\usepackage{subcaption}
\usepackage{arydshln}
\usepackage{makecell}
\usepackage[inline]{enumitem}
\usepackage{enumitem}
\usepackage{bm}

\usepackage{newfloat}
\usepackage{listings}
\DeclareCaptionStyle{ruled}{labelfont=normalfont,labelsep=colon,strut=off} 
\floatstyle{ruled}
\newfloat{listing}{tb}{lst}{}
\floatname{listing}{Listing}

\title{Black-box Model Merging for Language-Model-as-a-Service \\with Massive Model Repositories}
\author {
    Shilian Chen\textsuperscript{\rm 1},
    Jie Zhou\textsuperscript{\rm 1}\thanks{Corresponding author.},
    Tianyu Huai\textsuperscript{\rm 1},
    Yujiang Lu\textsuperscript{\rm 1},
    Junsong Li\textsuperscript{\rm 1},
    Bihao Zhan\textsuperscript{\rm 1},
    Qianjun Pan\textsuperscript{\rm 1},
    Yutao Yang\textsuperscript{\rm 1},
    Xin Li\textsuperscript{\rm 2},
    Qin Chen\textsuperscript{\rm 1},
    Hang Yan\textsuperscript{\rm 3},
    Liang He\textsuperscript{\rm 1}
}
\affiliations {
    \textsuperscript{\rm 1}School of Computer Science and Technology, East China Normal University, Shanghai\\
    \textsuperscript{\rm 2}Shanghai AI Laboratory, \textsuperscript{\rm 3}The Chinese University of HongKong
    \nocopyright
}

\begin{document}

\maketitle

\begin{abstract}
Model merging refers to the process of integrating multiple distinct models into a unified model that preserves and combines the strengths and capabilities of the individual models. Most existing approaches rely on task vectors to combine models, typically under the assumption that model parameters are accessible. However, for extremely large language models (LLMs) such as GPT-4, which are often provided solely as black-box services through API interfaces (Language-Model-as-a-Service), model weights are not available to end users. This presents a significant challenge, which we refer to as black-box model merging (\texttt{BMM}) with massive LLMs.
To address this challenge, we propose a derivative-free optimization framework based on the evolutionary algorithm (\texttt{Evo-Merging}) that enables effective model merging using only inference-time API queries. Our method consists of two key components: (1) sparsity-based denoising, designed to identify and filter out irrelevant or redundant information across models, and (2) sign-aware scaling, which dynamically computes optimal combination weights for the relevant models based on their performance. 
We also provide a formal justification, along with a theoretical analysis, for our asymmetric sparsification.
Extensive experimental evaluations demonstrate that our approach achieves state-of-the-art results on a range of tasks, significantly outperforming existing strong baselines \footnote{We will release the codes, datasets, and models on GitHub.}.
\end{abstract}

\section{Introduction}
Model merging \cite{yang2024model,akiba2025evolutionary}, also known as model fusion, represents a transformative technique within the machine learning landscape. It offers an efficient and resource-conscious approach to enhance the capabilities of artificial intelligence models, particularly those of substantial scale. Unlike conventional methods that typically necessitate extensive data collection or computationally intensive retraining processes, model merging directly integrates the parameters of multiple pre-trained or fine-tuned models into a single, cohesive unit. This methodology is rapidly gaining prominence across diverse scientific and engineering domains, especially with the widespread adoption and development of large-scale foundation models.

\begin{figure}[t!]
    \centering
    \includegraphics[width=0.46\textwidth]{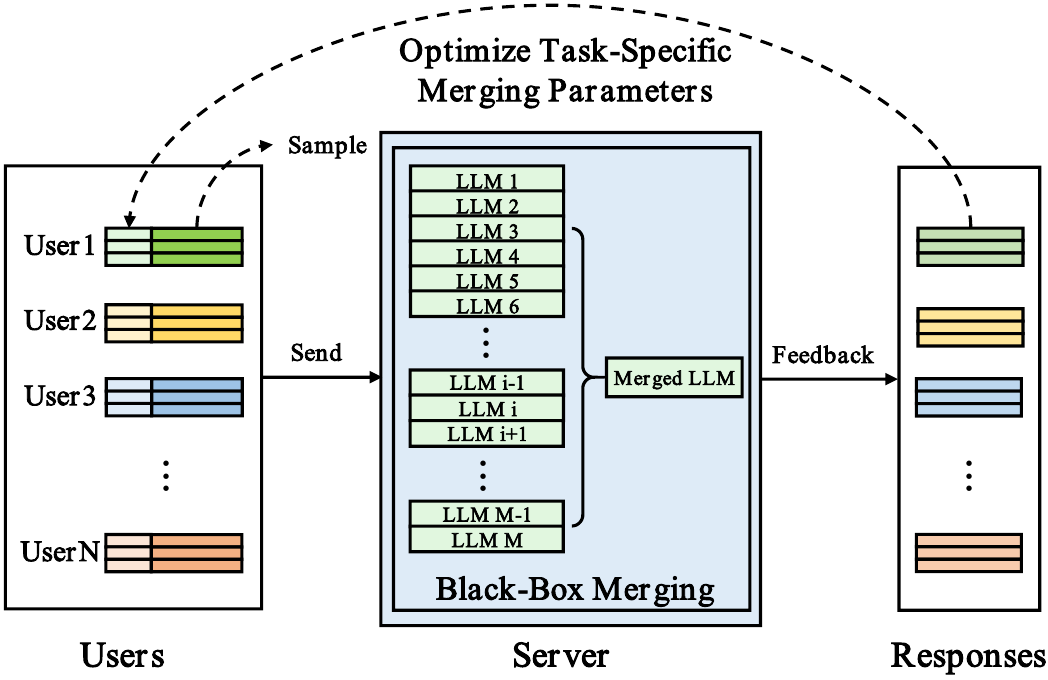}
    \vspace{-1mm}
    \caption{The process of Black-box Model Merging for Language-Model-as-a-Service.}
    \label{fig:intro}
\vspace{-4mm}
\end{figure}

Researchers have explored model merging techniques that combine the weights of several fine-tuned models at the parameter level, thereby creating a generalized or task-adapted model. Typical approaches fall into two categories. The first involves linear parameter interpolation \cite{matena2022merging}, which averages model weights with task-specific coefficients. 
The second is task-vector-based large-scale merging approaches (e.g., EMR-Merging \cite{huang2024emr}, Twin-Merging \cite{lu2024twin}, TIES-Merging \cite{yadav2023ties}, DARE \cite{yu2024language}), which aim to retain general multi-task capabilities across a small number of models, under controlled noise levels. 


Existing model merging methods are typically designed based on direct access to model parameters, making them user-unfriendly in practice. Nowadays, many large language models (LLMs) are deployed as services under the ``language-models-as-a-service'' paradigm \cite{sun2022black}, where the underlying parameters are inaccessible to end users. Moreover, these services are primarily used by non-expert users who lack deep knowledge of computer algorithms (They do not know which model to choose and how to merge the models), further highlighting the need for more accessible and parameter-free merging approaches (See Figure \ref{fig:intro}), defined as \textbf{black-box model merging (\texttt{BMM})}. 
Furthermore, \texttt{BMM} faces additional challenges, including the presence of irrelevant and potentially harmful information within these large models, which can degrade performance or lead to unintended behaviors. As a result, developing effective and safe merging strategies for such models requires innovative approaches that do not depend on internal parameter access and can filter out undesirable content while preserving useful knowledge.

To tackle these challenges, we propose \texttt{Evo-Merging}, a derivative-free framework for \texttt{BMM}. 
Specifically, we optimize the parameters without accessing the model weights based on feedback from the API using an evolutionary algorithm.
First, we design a sparsity-based denoising module to filter out irrelevant parameters, since not all model parameters contribute meaningfully to the current task.
Subsequently, we propose a sign-aware scaling module to adjust parameter magnitudes, as the significance of individual models differs for the target task. 
We also conduct a theoretical analysis to prove the effectiveness of our proposed model. 
\texttt{Evo-Merging} is designed to combine 100+ fine-tuned models with cross-task/domain knowledge. 
The fused model retains strong performance under both in-domain and out-of-domain settings. 

Our main contributions are as follows:

\begin{itemize}[leftmargin=*, align=left]
    \item We formalize a black-box model merging (\texttt{BMM}) task for language-model-as-a-service, where models are merged via API interfaces without access to their parameters.
    \item We develop an evolutionary algorithm-based framework, named \texttt{Evo-Merging}, to indicate the key knowledge from massive models for BMM. Specifically, we first propose a sparsity-based denoising module to eliminate irrelevant parameters, followed by a sign-aware scaling module that adaptively rescales model contributions based on task relevance.
    \item We demonstrate that \texttt{Evo-Merging} can effectively handles merging over 100 models while avoiding performance degradation, indicating its strong generalization and knowledge reuse abilities.
\end{itemize}

\section{Related Work}
Model merging is a pivotal research area for creating powerful, multi-talented LLMs from specialized models, as surveyed in recent work \cite{yang2024model}. Foundational methods often operate on task vectors, representing parameter-space differences. For instance, Task Arithmetic \cite{ilharco2023editing} showed that simple vector operations could combine skills, while Fisher-weighted averaging \cite{matena2022merging} proposed a more principled weighting scheme. Building on these ideas, a variety of advanced ``white-box'' techniques have emerged. TIES-Merging \cite{yadav2023ties} focuses on resolving sign-based conflicts to mitigate interference. DARE \cite{yu2024language} prunes and rescales task vectors to absorb abilities from homologous models. Other methods like EMR-Merging \cite{huang2024emr}, Twin-Merging \cite{lu2024twin}, Della-Merging \cite{deep2024dellamergingreducinginterferencemodel}, and Breadcrumbs \cite{davari2024modelbreadcrumbsscalingmultitask} explore high-rank fusion, dynamic integration, magnitude-based sampling, and sparse masks, respectively.

The domain of LoRa merging has also seen a surge of innovation. LoRaHub \cite{huang2024LoRaHubefficientcrosstaskgeneralization} and LoRa-Flow \cite{wang2024LoRaflowdynamicLoRafusion} propose dynamic composition or fusion of LoRa modules, where merging weights are adjusted per-input at inference time. Others focus on more sophisticated static merging, such as using SVD for a principled fusion \cite{stoica2024modelmergingsvdtie}, performing adaptive merging with parameter pruning \cite{miyano2025adaptiveLoRamergeparameter}, or even exploring extreme modularity through rank-wise clustering of LoRa components \cite{zhao2024mergingLoRaslikeplaying}.

However, existing methods typically rely on direct parameter access, rendering them incompatible with the Black-box Model Merging (\texttt{BMM}) paradigm under language-modal-as-a-service settings \cite{sun2022black}. Moreover, these approaches tend to suffer performance degradation in the presence of irrelevant domain knowledge, especially when scaling to massive tasks. 
Unlike prior work, which either assumes full model access or lacks effective mechanisms for denoising a large pool of models in black-box environments, our \texttt{Evo-Merging} is specifically designed for this challenging setting. Built upon the derivative-free CMA-ES algorithm \cite{hansen1996adapting}, \texttt{Evo-Merging} operates without gradients, making it particularly well-suited for \texttt{BMM}, where preserving user data privacy and ensuring high performance are of utmost importance.

\begin{figure*}[t!]
    \centering
    \includegraphics[width=0.98\textwidth]{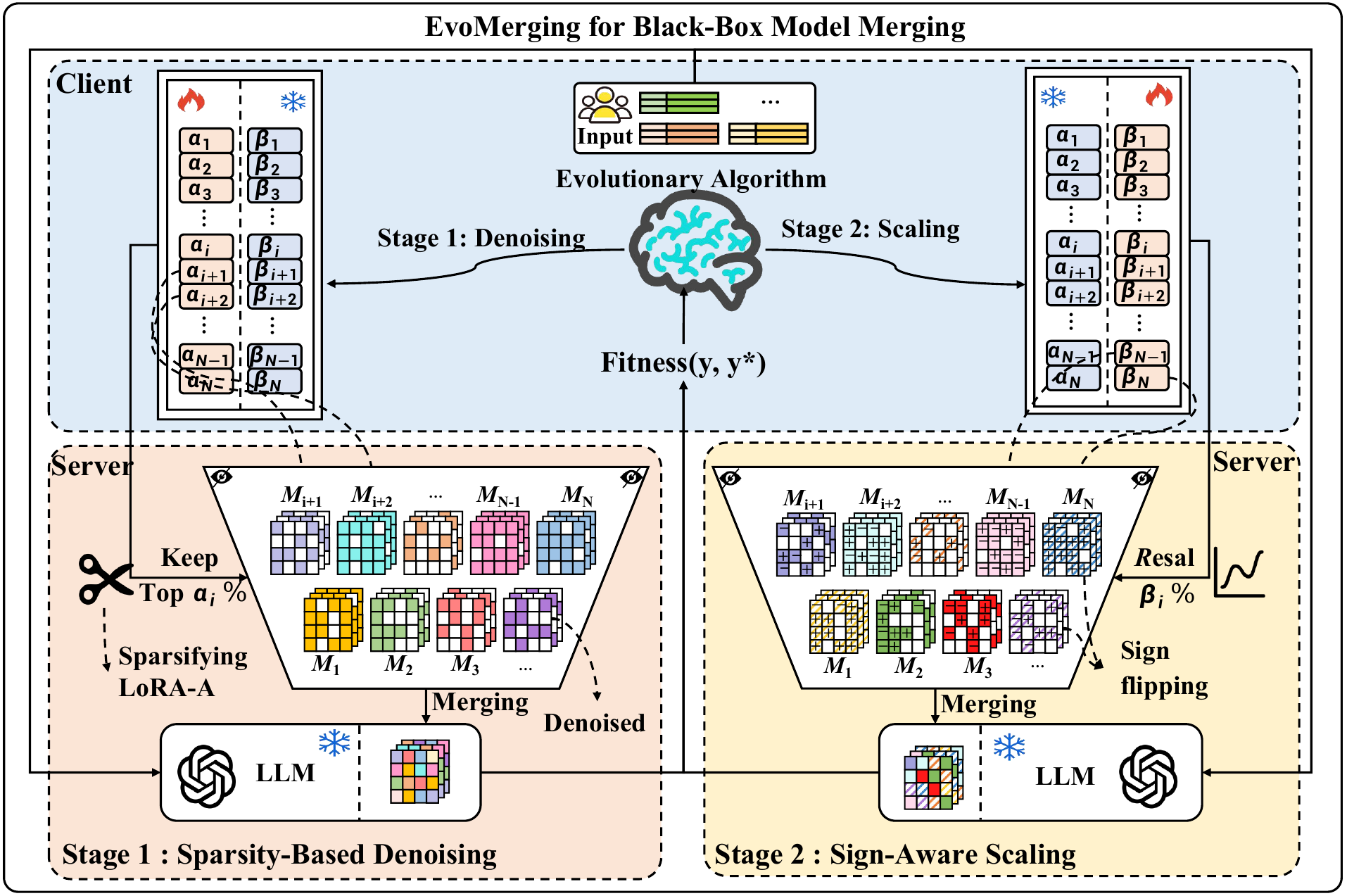}
    \vspace{-1mm}
    \caption{The Evo-Merging framework of Black-box Model Merging for Language-Model-as-a-Service. Users send inputs to the server, and an evolutionary algorithm is used to optimize the parameters ($\mathbf{\alpha}$ and $\mathbf{\beta}$) based on feedback from API interfaces. }
    \label{fig:framework}
\vspace{-4mm}
\end{figure*}

\section{Our Methodology}
\label{sec:methodology}
We present \texttt{Evo-Merging} (Figure \ref{fig:framework}), a novel black-box model merging (\texttt{BMM}) framework designed for the language-model-as-a-service paradigm \cite{sun2022black}, where internal model parameters are inaccessible. Our approach operates without gradients, instead leveraging an evolutionary algorithm guided by API-based feedback. The framework comprises two key components: a sparsity-based denoising module that filters out irrelevant model contributions and a sign-aware scaling module that adaptively rescales model influences. 
We also provide a comprehensive theoretical and empirical analysis, supported by formal justification, to demonstrate the effectiveness of our \texttt{Evo-Merging}.

\subsection{Preliminaries}
\label{sec:preliminaries}

\paragraph{Task Definition.}
Let $\mathcal{M} = \{M_1, M_2, \dots, M_N\}$ denote a collection of $N$ pre-trained or fine-tuned black-box language models, they are accessible only through an API under the ``language models as a service'' paradigm \cite{sun2022black}, where internal parameters are not available to end users. 
Given a target task $\mathcal{T}$ defined over an input space $\mathcal{X}$ and output space $\mathcal{Y}$, the objective of \texttt{BMM} is to construct a fused model $M^*$ that maps inputs $x \in \mathcal{X}$ to outputs $y \in \mathcal{Y}$ by effectively aggregating knowledge from all models in $\mathcal{M}$, without direct access to their internal parameters.

Formally, we aim to learn a merging function $\Phi(\cdot)$:
\begin{equation}
    M^*(x) = \Phi(M_1(x), M_2(x), \dots, M_N(x); \theta),
\end{equation}
where $\theta$ is the learnable parameter, $M^*$ achieves strong performance on $\mathcal{T}$, generalizes well to out-of-domain examples, and remains robust to irrelevant or harmful information present in individual source models.

\paragraph{Task Vectors and Low-Rank Representation.}
A task vector, $\Delta W = W_{\text{ft}} - W_{\text{pre}}$, represents the parameter-space difference between a fine-tuned model and its pre-trained base, encapsulating task-specific knowledge \cite{ilharco2023editing}. However, the high dimensionality of $\Delta W$ makes its direct manipulation computationally prohibitive.

The Low-Rank Adaptation (LoRa) method \cite{hu2021LoRa} addresses this by positing that $\Delta W$ has a low intrinsic rank. Consequently, the weight change for a layer can be approximated by a low-rank product:
\begin{equation}
    \Delta W^{(l)} \approx B^{(l)} A^{(l)},
    \label{eq:LoRa_approx}
\end{equation}
where for a layer mapping $\mathbb{R}^k \to \mathbb{R}^d$, the matrices $A \in \mathbb{R}^{r \times k}$ and $B \in \mathbb{R}^{d \times r}$ have a small rank $r \ll \min(d, k)$. This factorization allows fusion operations to be performed on the compact matrices $\{A_i, B_i\}$ instead of the full-parameter $\Delta W$, drastically reducing computational complexity.

\subsection{Evo-Merging}
\label{sec:two_stage_optimization}
Evo-Merging employs a two-stage process based on the general formulation for the merged LoRa pair, $(\mathcal{A}_m, \mathcal{B}_m)$:
\begin{align}
    \mathcal{A}_m &= \sum_{i=1}^{N} w_i \cdot \mathbf{S}_{\alpha_i}(A_i) \label{eq:general_form_A} \\
    \mathcal{B}_m &= \sum_{i=1}^{N} w_i \cdot B_i \label{eq:general_form_B}
\end{align}

\paragraph{Stage 1: Sparsity-Based Denoising.}

In the initial stage, this operator creates a sparse matrix $A_i' = \mathbf{S}_{\alpha_i}(A_i)$ by retaining the top $\alpha_i\%$ of parameters with the highest absolute values in $A_i$ and setting all other elements to zero. We provide theoretical and empirical analysis in the following section to justify sparsifying $A$ —rather than $B$ or both.

With the merge weights temporarily fixed to a uniform distribution ($w_i = 1/N$), our objective is to determine the optimal set of sparsity ratios, denoted by the vector $\bm{\alpha}^* = [\alpha_1^*, \dots, \alpha_N^*]^\top$. This is framed as an optimization problem solved using the CMA-ES~\cite{hansen1996adapting}:
\begin{equation}
\begin{aligned}
    \underset{\bm{\alpha}}{\text{min}} \quad & \mathcal{F}(\bm{\alpha}) = \mathcal{L}_{\text{CE}}(\mathcal{D}_{\text{val}}, \mathcal{M}_{\text{adapter}}(\bm{\alpha})) \\
    & \qquad\qquad + \lambda_1 ||\bm{\alpha}||_1 \\
    \text{subject to} \quad & 0 \le \alpha_i \le 1, \quad \forall i=1, \dots, N
\end{aligned}
\label{eq:fitness_stage1}
\end{equation}
Here, $\mathcal{M}_{\text{adapter}}(\bm{\alpha})$ represents the complete adapter constructed from the pre-merged pair $(A_{\text{pm}}(\bm{\alpha}), B_{\text{pm}})$, where $A_{\text{pm}}(\bm{\alpha}) = \sum_{i=1}^{N} w_i \cdot \mathbf{S}_{\alpha_i}(A_i)$ and  $B_{\text{pm}}= \sum_{i=1}^{N} w_i \cdot \mathbf (B_i)$  . The $L_1$ regularization term, $||\bm{\alpha}||_1$, encourages sparsity among the selected ratios themselves. This compels the evolutionary search to effectively prune entire adapters that are deemed irrelevant or excessively noisy by driving their corresponding sparsity ratios $\alpha_i$ towards zero.

\paragraph{Stage 2: Sign-Aware Scaling.}
The second stage refines the merge by optimizing the contribution of each (now sparsified) adapter. We freeze the optimized sparsity structure $\bm{\alpha}^*$ and optimize a new scaling vector $\bm{\beta} = [\beta_1, \dots, \beta_N]^\top$, setting the weights to $w_i = \beta_i$. The sign of $\beta_i$ provides a powerful mechanism for conflict resolution: $\beta_i > 0$ enhances synergy, while $\beta_i < 0$ subtracts conflicting knowledge. 
We treat the entire task vector  $\Delta W$ as each model’s knowledge carrier and dynamically adjust the magnitude of  $\beta$ to attenuate or amplify the fused model’s knowledge and capabilities. This dynamic scaling activates out‑of‑domain generalization, unlocks cross‑domain knowledge, and elicits emergent capabilities when tackling entirely new tasks. 
The effect of these techniques is analyzed in detail in Section~\ref{sec: Effect of Sparsity-Based Denoising and Sign-Aware Scaling.} 
We again use CMA-ES~\cite{hansen1996adapting} to find the optimal $\bm{\beta}$ by minimizing:
\begin{equation}
\begin{aligned}
    \min_{\bm{\beta}} \mathcal{G}(\bm{\beta}) = \mathcal{L}_{\text{CE}}(\mathcal{D}_{\text{val}}, \mathcal{M}_{\text{adapter}}(\bm{\beta})) + \lambda_2 ||\bm{\beta}||_1
\end{aligned}
\label{eq:fitness_stage2}
\end{equation}
The $L_1$ regularization prevents certain adapters from being overly amplified and suppresses noise from irrelevant adapters toward zero.

\subsection{Theoretical and Empirical Analysis}
\label{sec:Theoretical_and_Experimental_Analysis}

In this section, our goal is to justify our core design choice: asymmetric sparsification. We provide a robust theoretical and empirical argument for why we exclusively prune the LoRa-A matrix and not the B matrix or the full task vector. Our central claim is that this is not just a computationally convenient choice, but a principled and more effective strategy for model denoising.

\paragraph{Conceptual Hypothesis.}


Our strategy is rooted in a conceptual model where the LoRa matrices serve distinct, complementary roles: The A matrix acts as a "concept extraction" layer, and the B matrix as a "knowledge mapping" layer. This functional view is conceptually aligned with the subspace analysis by \cite{si2024parameterefficientfinetuningstanding}, which mathematically frames the columns of matrix A as providing ``new bases" to adapt the model to a new task. In our model, matrix A therefore learns to identify the salient concepts (``what to look for"), while matrix B maps these concepts into the output space ("how to act on it").

Under this model, task-related ``noise''---such as learning spurious or irrelevant correlations from the training data---is far more likely to manifest as the extraction of non-essential concepts in matrix A. Therefore, sparsifying matrix A should be the most direct way to denoise the task vector by eliminating these irrelevant concepts at their source.

\paragraph{Formal Justification.}
While the conceptual model provides the intuition, we can formalize our approach to show it is mathematically sound.
\begin{itemize}[leftmargin=*]
    \item \textbf{The Ideal (But Intractable) Objective.}
Ideally, we want to find a sparse version of the task vector, let's call it $\Delta W'$, that is as close as possible to a hypothetical, perfectly denoised task vector $\Delta W^*$. The objective would be:
\begin{equation}
    \min_{\Delta W'} \| \Delta W' - \Delta W^* \|_F
\end{equation}
However, $\Delta W^*$ is unknown and finding it directly is intractable due to the immense search space.
    \item \textbf{The Practical Proxy Objective.}
Instead, we adopt a practical proxy: we find the optimal sparsity ratios, $\bm{\alpha}$, for matrix A by minimizing the validation loss on a hold-out set:
\begin{equation}
    \bm{\alpha}^* = \underset{\bm{\alpha}}{\text{argmin}} \mathcal{L}(B \cdot \mathbf{S}_{\bm{\alpha}}(A), \mathcal{D}_{\text{val}})
\end{equation}
where $\mathbf{S}_{\bm{\alpha}}(A)$ is the sparsified A matrix.
\end{itemize}

\begin{figure}[t!]
    \centering
    \includegraphics[width=1\linewidth]{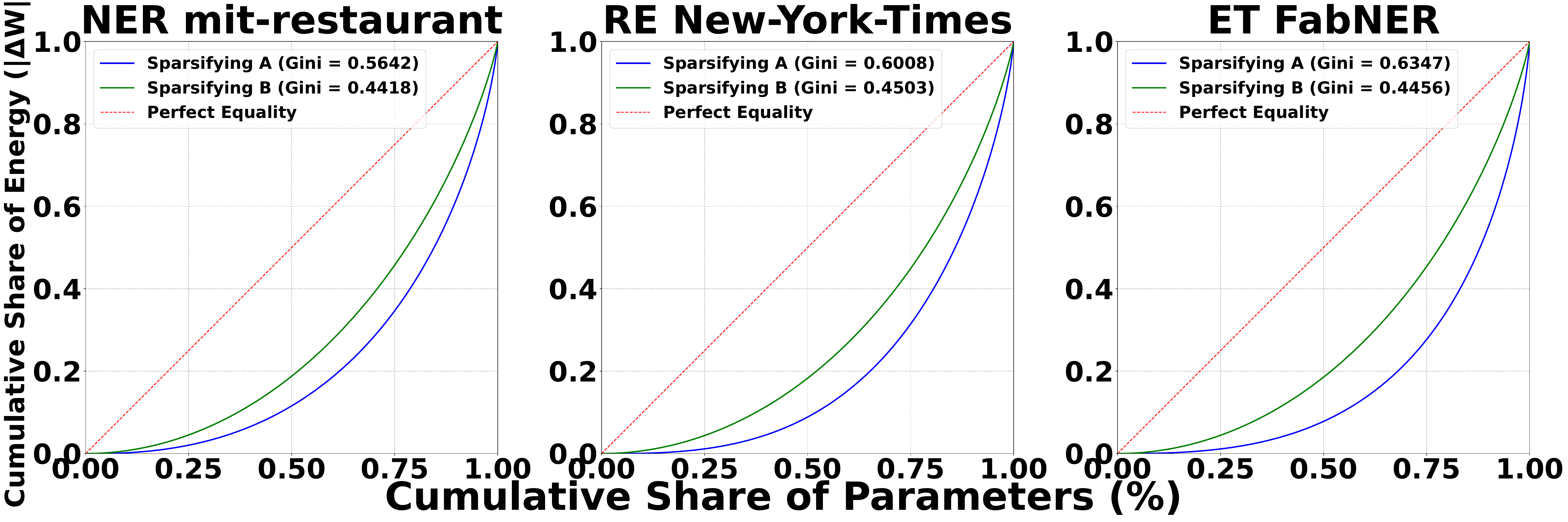}
    \vspace{-5mm}
    \caption{Lorenz curve of the final $\Delta W$ for Sparsifying A vs. Sparsifying B. 
    }
    \label{fig:gini_curve}
    \vspace{-2mm}
\end{figure}

\begin{figure}[t!]
    \centering
    \includegraphics[width=0.8\linewidth]{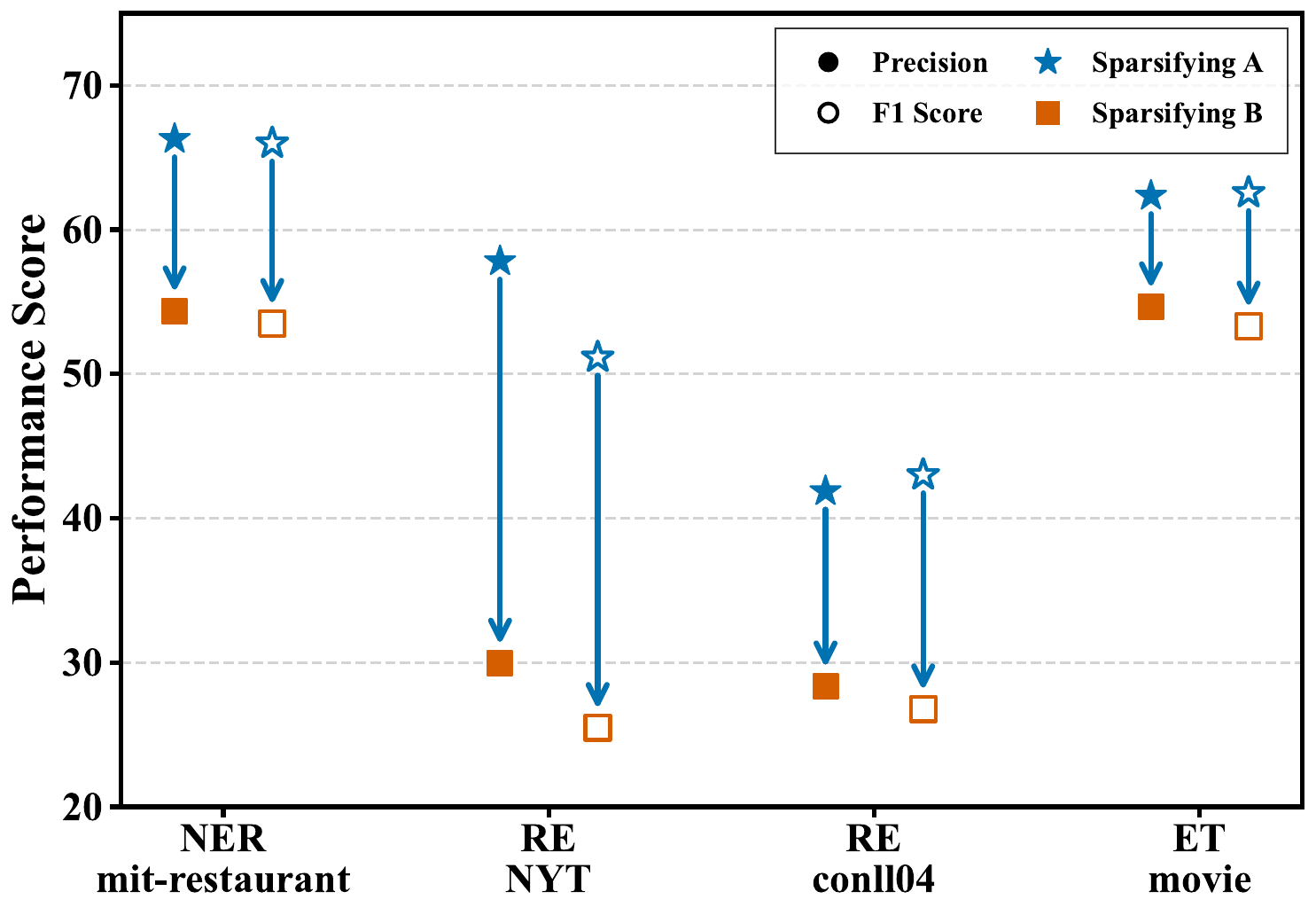}
    \vspace{-1mm}
    \caption{Performance degradation when applying sparsification with Sparsifying A and Sparsifying B.}
    \label{fig:perf_drop}
     \vspace{-2mm}
\end{figure}

\begin{table*}[t!]
\small
\vspace{-2mm}
\centering
\setlength{\tabcolsep}{3.5pt}
\begin{tabular}{l cc cc cc cc cc cc c c}
\toprule 
\multirow{3}{*}{{Method}}  & \multicolumn{4}{c}{{NER}} & \multicolumn{4}{c}{{RE}} & \multicolumn{4}{c}{{ET}} & \multicolumn{2}{c}{\multirow{2}{*}{{Avg}}} \\ 
\cmidrule(lr){2-5} \cmidrule(lr){6-9} \cmidrule(lr){10-13} 
 & \multicolumn{2}{c}{{Mit-Restaurant}} & \multicolumn{2}{c}{{FindVehicle}} & \multicolumn{2}{c}{{New York Time}} & \multicolumn{2}{c}{{Conll04}} & \multicolumn{2}{c}{{Mit-Movie}} & \multicolumn{2}{c}{{FabNER}} & & \\
\cmidrule(lr){2-3} \cmidrule(lr){4-5} \cmidrule(lr){6-7} \cmidrule(lr){8-9} \cmidrule(lr){10-11} \cmidrule(lr){12-13} \cmidrule(lr){14-15} 
  & Prec & F1 & Prec & F1 & Prec & F1 & Prec & F1 & Prec & F1 & Prec & F1 & Prec & F1 \\
\midrule 
TIES            & 21.40 & 24.22 & 29.87 & 29.45 & 12.55 & 7.57  & 20.78 & 19.05 & 48.67 & 51.12 & 15.91 & 11.42 & 24.86 & 23.81 \\
DARE (TIES+)    & 10.88 & 14.31 & 24.13 & 23.59 & 3.69  & 2.53  & 10.85 & 8.08  & 44.57 & 48.60 & 15.44 & 10.71 & 18.26 & 17.97 \\
Task Arithmetic & 12.02 & 11.57 & 14.29 & 14.96 & 15.24 & 16.04 & 11.44 & 12.03 & 36.70 & 41.36 & 9.90  & 9.68  & 16.60 & 17.61 \\
Breadcrumbs      & 38.99 & 34.06 & 47.41 & 38.42 & 34.05 & 28.03 & 26.30 & 24.56 & 49.73 & 51.60 & 20.53 & 16.84 & 36.17 & 32.25 \\
Della           & 29.69 & 26.06 & 16.66 & 17.04 & 3.94  & 3.25  & 19.91 & 17.97 & 43.85 & 47.68 & 14.41 & 12.51 & 21.41 & 20.75 \\
EMR-Merging     & 31.89 & 27.46 & 46.92 & 42.25 & 32.49 & 28.94 & 16.45 & 17.10 & 46.57 & 47.35 & 19.26 & 17.86 & 32.26 & 30.16 \\
LoRaHub         & 65.05 & 63.08 & 31.81 & 30.75 & 54.23 & 48.61 & 21.30 & 21.77 & 41.99 & 41.70 & 41.94 & 39.74 & 42.72 & 40.94 \\
Best Adapter    & 37.71 & 35.99 & 39.11 & 38.49 & 45.55 & 38.77 & 11.47 & 11.33 & 46.76 & 46.16 & \textbf{48.45} & \textbf{47.16} & 38.18 & 36.32 \\
LoRa            & 28.18 & 29.65 & \textbf{52.19} & \textbf{52.34} & 52.05 & 44.66 & \textbf{45.42} & 41.60 & 25.20 & 25.38 & 18.74 & 20.01 & 36.96 & 35.61 \\
(IA)$^3$        & 36.18 & 41.64 & 24.98 & 26.05 & 30.54 & 30.70 & 17.01 & 19.97 & 38.80 & 42.17 & 15.38 & 16.70 & 27.15 & 29.54 \\
\midrule 
\textbf{\texttt{Evo-Merging}} & \textbf{66.29} & \textbf{65.99} & 49.61 & 47.64 & \textbf{57.81} & \textbf{51.15} & 41.87 & \textbf{43.00} & \textbf{64.11} & \textbf{62.55} & 43.10 & 42.47 & \textbf{53.80} & \textbf{52.13} \\
\bottomrule
\end{tabular}
\caption{Main results of our methods and the selected baselines over out-of-domain (OOD) tasks.}
\label{tab:method_performance_ood}
\vspace{-2mm}
\end{table*}

\paragraph{Why This Proxy is Sound.}
Our choice to manipulate $A$ is justified because its impact on the final task vector is well-behaved and controllable. The error introduced by sparsifying $A$ is bounded:
\begin{equation}
\begin{aligned}
       \| \Delta W - \Delta W'(\bm{\alpha}) \|_F & = \| B \cdot A - B \cdot \mathbf{S}_{\bm{\alpha}}(A) \|_F \\
       & \leq \| B \|_F \cdot \| A - \mathbf{S}_{\bm{\alpha}}(A) \|_F 
\end{aligned}
    \label{eq:error_bound}
\end{equation}

This inequality is the core of our formal argument. It demonstrates that the total approximation error in the task vector $\Delta W$ is directly controlled by how well the sparsified matrix $\mathbf{S}_{\bm{\alpha}}(A)$ approximates the original matrix $A$. By performing magnitude-based pruning on $A$, we are directly minimizing the term $\| A - \mathbf{S}_{\bm{\alpha}}(A) \|_F$, which in turn constrains the overall error. This provides a stable, principled mechanism for denoising.

Finally, we connect our practical objective (minimizing validation loss) to the ideal one (getting close to $\Delta W^*$) with the following standard machine learning assumptions:
\begin{itemize}[leftmargin=*]
    \item \textbf{Optimality of Ideal Solution:} The ideal vector $\Delta W^*$ is, by definition, the minimizer of the true, unobserved loss: $\Delta W^* = \underset{\Delta W'}{\text{argmin}} \mathcal{L}(\Delta W', \mathcal{D}_{\text{true}})$.
    \item \textbf{Proxy Quality:} The validation loss is a smooth and faithful estimator of the true loss.
\end{itemize}
This creates a logical chain: minimizing the validation loss serves as a proxy for minimizing the true loss, which ideally leads to a solution close to the optimal sparse vector $\Delta W^*$. We use a derivative-free optimizer (CMA-ES) to find a near-optimal $\bm{\alpha}$.

\paragraph{Empirical Validation.}
The conceptual hypothesis and formal justification lead to a clear, testable prediction: If matrix A is indeed the primary locus of learnable concepts and noise, then sparsifying it should be more effective and safer than sparsifying matrix B.
We test this by comparing our proposed asymmetric approach (Sparsifying A) against the alternative (Sparsifying B). Our empirical results in Figures~\ref{fig:gini_curve} and~\ref{fig:perf_drop} confirm our prediction: 
1) It results in a final task vector $\Delta W$ with a higher Gini coefficient, indicating a sparser and more concentrated representation of knowledge;  
2) The ``Sparsifying A'' strategy consistently retains higher performance on downstream tasks after merging.
This provides strong empirical evidence that matrix $A$ contains a higher density of both task-specific knowledge and noise, making it the right target for sparsification.
This finding aligns with the conclusions from \cite{benedek2024priLoRaprunedrankincreasinglowrank} which also examined that pruning matrix A is a safer and more effective strategy than targeting matrix B.

\section{Experiments}

This section evaluates our proposed Black-Box Merging (\texttt{BMM}) framework, \texttt{Evo-Merging}. We benchmark it against a suite of baselines and perform in-depth ablation studies to validate our design choices. Evo-Merging 's implementation details are in appendix ~\ref{sec:Evo-Merging Implement Detail} and ~\ref{alg: Algorithm flow of Evo-Merging}.

\subsection{Experimental Setups}
\subsubsection{Datasets and Metrics.}
To demonstrate our model's efficacy, we evaluate its performance on large-scale in-domain and out-of-domain tasks. Our framework uses the Llama 3.1 Instruct model~\cite{meta2024llama3.1} and the InstructUIE benchmark~\cite{wang2023instructuiemultitaskinstructiontuning}—chosen for its generative complexity—to fine-tune over 100 models on information extraction tasks (e.g., NER, RE, and EE). We evaluate all methods using Macro Precision (Prec) and Macro F1-Score (F1). Further details on the datasets and metrics are in Appendix~\ref{sec: Dataset Details and Evalution}.

\begin{table*}[t!]
\footnotesize  
\centering
\vspace{-2mm}
\setlength{\tabcolsep}{2.0pt} 
\begin{tabular}{l *{9}{cc}}
\toprule 
\multirow{2}{*}{\textbf{Method}}
& \multicolumn{2}{c}{ACE\_2005} & \multicolumn{2}{c}{CrossNER\_AI} & \multicolumn{2}{c}{CrossNER\_Lit} & \multicolumn{2}{c}{FabNER} & \multicolumn{2}{c}{Mit-Movie} & \multicolumn{2}{c}{MultiNERD} & \multicolumn{2}{c}{TweetNER} & \multicolumn{2}{c}{WikiANN} & \multicolumn{2}{c}{Avg} \\
\cmidrule(lr){2-3} \cmidrule(lr){4-5} \cmidrule(lr){6-7} \cmidrule(lr){8-9} \cmidrule(lr){10-11} \cmidrule(lr){12-13} \cmidrule(lr){14-15} \cmidrule(lr){16-17} \cmidrule(lr){18-19} 
& Prec & F1 & Prec & F1 & Prec & F1 & Prec & F1 & Prec & F1 & Prec & F1 & Prec & F1 & Prec & F1 & Prec & F1 \\
\midrule
Individual model & 70.45 & 69.37 & 9.51 & 21.23 & 20.05 & 11.83 & 48.61 & 47.09 & 74.27 & 74.38 & 86.15 & 85.55 & 62.24 & 58.26 & 81.33 & 80.82 & 56.58 & 56.07 \\
\hline 
TIES & 16.16 & 13.11 & 18.99 & 18.93 & 27.27 & 26.47 & 21.02 & 18.91 & \textbf{54.77} & 51.24 & 40.70 & 48.99 & 34.05 & 33.96 & 55.30 & 56.47 & 33.53 & 33.51 \\
DARE & 12.72 & 13.50 & 20.40 & 18.42 & 25.97 & 25.13 & 19.11 & 18.47 & 53.12 & 51.67 & 51.58 & 47.14 & 32.97 & 32.91 & 54.13 & 55.58 & 33.75 & 32.85 \\
DELLA & 15.73 & 12.74 & 17.72 & 18.02 & 25.52 & 25.16 & 19.98 & 18.55 & 54.65 & \textbf{51.94} & 38.69 & 47.11 & 31.99 & 32.30 & 54.61 & 56.15 & 32.36 & 32.75 \\
Breadcrumbs & 19.74 & 15.34 & 21.33 & 20.03 & 31.87 & 29.00 & 21.18 & 17.43 & 51.46 & 46.91 & 48.44 & 55.53 & 37.26 & 33.65 & 56.56 & 56.82 & 35.98 & 34.34 \\
Task Arithmetic & 17.37 & 13.40 & 20.12 & 18.90 & 30.55 & 28.38 & 20.86 & 17.63 & 53.59 & 49.27 & 45.43 & 53.04 & 35.60 & 33.47 & 56.33 & 56.93 & 34.98 & 33.88 \\
LoRaHub & \textbf{28.36} & \textbf{25.18} & 33.67 & 33.69 & 33.23 & 31.51 & 26.04 & 22.32 & 38.90 & 36.48 & 36.61 & 43.89 & 26.99 & 30.04 & 53.78 & 55.32 & 34.70 & 34.80 \\
EMR-Merging & 19.08 & 15.10 & 16.99 & 16.16 & 26.09 & 23.12 & 23.70 & 19.76 & 52.09 & 46.67 & 46.65 & 54.16 & \textbf{37.91} & 34.76 & \textbf{59.01} & \textbf{59.49} & 35.19 & 33.65 \\
\midrule 
\textbf{\texttt{Evo-Merging}} & 26.31 & 23.46 & \textbf{34.18} & \textbf{34.61} & \textbf{33.95} & \textbf{32.32} & \textbf{27.15} & \textbf{23.27} & 43.58 & 39.94 & \textbf{52.82} & \textbf{56.59} & 36.34 & \textbf{35.08} & 58.40 & 58.98 & \textbf{39.09} & \textbf{38.03} \\

\bottomrule 
\end{tabular}
\caption{Main results of our methods and selected baselines over in-domain tasks.}
\label{tab:method_performance_indomain}
\end{table*}

\begin{table*}[t]
\small
\centering
\vspace{-2mm}
\setlength{\tabcolsep}{0.5mm} 
\begin{tabular}{lcccccccccccc cc} 
\toprule
\multirow{2}{*}{Method}  & \multicolumn{2}{c}{NER\_Mit-Rest} & \multicolumn{2}{c}{NER\_FindVeh} & \multicolumn{2}{c}{RE\_NYT} & \multicolumn{2}{c}{RE\_Conll04} & \multicolumn{2}{c}{ET\_Mit-Movie} & \multicolumn{2}{c}{ET\_FabNER} & \multicolumn{2}{c}{Avg} \\
\cmidrule(lr){2-3} \cmidrule(lr){4-5} \cmidrule(lr){6-7} \cmidrule(lr){8-9} \cmidrule(lr){10-11} \cmidrule(lr){12-13} \cmidrule(lr){14-15}
 & Prec & F1 & Prec & F1 & Prec & F1 & Prec & F1 & Prec & F1 & Prec & F1 & Prec & F1 \\
\midrule
\textbf{\texttt{Evo-Merging}} & \textbf{66.29} & \textbf{65.99} & \textbf{49.61} & \textbf{47.64} & \textbf{57.81} & \textbf{51.15} & \textbf{41.87} & \textbf{43.00} & \textbf{64.11} & \textbf{62.55} & \textbf{43.10} & \textbf{42.47} & \textbf{53.80} & \textbf{52.13} \\
\midrule
w/o Denoising \& Scaling & 28.01 & 33.10 & 44.14 & 32.68 & 2.35 & 1.94 & 7.36 & 5.37 & 43.70 & 39.67 & 14.10 & 11.75 & 23.28 [$\downarrow$ 30.5] & 20.75 [$\downarrow$ 31.4] \\

w/o Denoising & 65.05 & 63.08 & 31.81 & 30.75 & 54.23 & 48.61 & 21.30 & 21.77 & 41.99 & 41.70 & 41.94 & 39.74 & 42.72 [$\downarrow$ 11.1] & 40.94 [$\downarrow$ 11.2] \\

w/o Scaling & 39.90 & 37.41 & 48.78 & 41.35 & 2.27 & 2.23 & 7.14 & 6.38 & 46.10 & 46.52 & 14.10 & 11.75 & 26.38 [$\downarrow$ 27.4] & 24.27 [$\downarrow$ 27.9] \\

w/o Sign-flipped & 42.04 & 37.70 & 46.61 & 41.07 & 45.45 & 38.38 & 35.21 & 32.91 & 62.19 & 60.47 & 34.77 & 29.80 & 44.38 [$\downarrow$ 9.4] & 40.06 [$\downarrow$ 12.1] \\
\bottomrule
\end{tabular}
\vspace{-2mm}
\caption{Ablation study of \texttt{Evo-Merging}'s components. [$\downarrow$] means the performance drop relative to our full model.}
\label{tab:detailed_ablation_final_v2}
\end{table*}

\subsubsection{Baselines.}

Our baselines are primarily divided into two categories: advanced model merging methods and setting-specific baselines. The first category includes a comprehensive suite of merging techniques, encompassing prominent methods like \textbf{TIES}~\cite{yadav2023ties}, \textbf{DARE}~\cite{yu2024languagemodelssupermario}, and \textbf{Task Arithmetic}~\cite{ilharco2022editing}, alongside contemporary approaches such as \textbf{DELLA}~\cite{deep2024dellamergingreducinginterferencemodel}, \textbf{EMR-Merging}~\cite{huang2024emr}, \textbf{Breadcrumbs}~\cite{davari2024modelbreadcrumbsscalingmultitask}, and \textbf{LoRaHub}~\cite{huang2024LoRaHubefficientcrosstaskgeneralization}. The second category consists of baselines for specific experimental settings: for out-of-domain setting, we include \textbf{LoRa}~\cite{hu2021LoRa}, \textbf{(IA)$^3$}~\cite{liu2022few}, and the \textbf{Best Adapter} (the top-performing source adapter on the validation set); for the in-domain setting, an \textbf{Individual Model} is used as a dedicated baseline. More details about the baseline  are provided in the Appendix~\ref{sec: BaseLine Details}.
\subsection{Main Results} 

\subsubsection{Out-of-domain Performance}
In the out-of-domain (OOD) setting, we fuse a pool of over 100 non-target models for an unseen task. 
To ensure a fair comparison, all baseline methods utilize the same validation set for merging, as detailed in Appendix \ref{sec: Pre-selection Strategy for Baselines in OOD Setting}.
From the results(Table~\ref{tab:method_performance_ood}), we observe the following findings.
\textbf{First}, \texttt{Evo-Merging} achieves over 10 points improvement in both precision and F1 score compared to the best baseline (53.80/52.13 vs. 42.72/40.94), even though the baselines have access to the models' parameters.
\textbf{Second}, \texttt{Evo-Merging} significantly outperforms the {Best Adapter} baseline (51.44\% vs. 31.10\% average F1). This result demonstrates that our framework effectively integrates complementary knowledge, rather than merely selecting the best model. In contrast, other methods are capped by the best single adapter's performance, revealing a flawed assumption that all source models are beneficial. Their inability to denoise and integrate knowledge from a large, noisy pool underscores the novelty and necessity of our approach (More analysis is shown in Appendix ~\ref{sec:more_analysis}).

\subsubsection{In-domain Performance}
In the standard in-domain setting, we fuse $N$ task-specific models into a single model that performs well on multiple tasks. 
As detailed in Table~\ref{tab:method_performance_indomain}, \texttt{Evo-Merging} sets a new state-of-the-art with a 38.03\% F1 score. This surpasses strong baselines like LoRaHub (34.80\% F1) and demonstrates superior conflict resolution. A performance gap to the individual model upper bound (56.07\% F1) persists, reflecting the core challenge in model merging: aligning conflicting label schemas. \texttt{Evo-Merging} addresses this more effectively than prior works by preserving task-specific knowledge while minimizing interference.

\subsection{Ablation Studies}

To isolate the contribution of each component in our framework, we conduct comprehensive ablation studies, with average results presented in Table~\ref{tab:detailed_ablation_final_v2}. 
The results clearly demonstrate the necessity of our two-stage approach. A naive average of all models (w/o Denoising \& Scaling) yields a mere 20.75\% average F1, confirming that simple fusion is ineffective. Removing only the denoising stage (w/o Denoising) improves the score to 40.94\%, but this remains far below the full model's 52.13\%, proving that denoising is a critical prerequisite. Conversely, removing the scaling stage (w/o Scaling) causes performance to collapse to 24.27\%, showing that a learned weighting scheme is crucial even for denoised models. Finally, disabling negative weights (w/o Sign-flipped) limits the F1 score to 40.06\%, underscoring the importance of actively subtracting conflicting knowledge for optimal results. 
Sparsity-based denoising and sign-aware scaling are not only individually essential but also highly synergistic. The denoising stage provides a clean knowledge base, which the scaling stage maximizes by resolving conflicts. Together, they produce a fusion effect superior to their individual components or naive fusion.

\subsection{Further Analysis}
\label{sec:Further Analysis}

\paragraph{Robustness Against Cross-task/domain Interference.}

To evaluate robustness, we create a noisy merging environment by adding five cross-task/domain models to an 8-task benchmark. As shown in Table~\ref{tab:drop_performance_direct}, conventional methods like DARE and DELLA suffer a catastrophic performance collapse (dropping over 20 F1 points). This failure occurs because these methods cannot filter noise from cross-task/domain and incorrectly assume all source models are beneficial. In stark contrast, \texttt{Evo-Merging} demonstrates exceptional resilience, maintaining the highest performance. More analysis about the results is shown in Appendix \ref{sec: More analyse about Evo-Merging on Muti-task merging}.

\begin{table}[t!]
\small
\centering
\vspace{-2mm}
\setlength{\tabcolsep}{0.8mm} 
\begin{tabular}{l  cc c cc c}
\toprule
\multirow{2}{*}{Method} 
  & \multicolumn{3}{c}{Precision (\%)} 
  & \multicolumn{3}{c}{F1 (\%)} \\ 
\cmidrule(lr){2-4} \cmidrule(lr){5-7}
 & 8-Task & 13-Task & Drop 
 & 8-Task & 13-Task & Drop \\
\midrule
TIES            & 33.53 & 12.82 & -20.71 & 33.51 & 12.05 & -21.46 \\
DARE            & 33.75 & 7.97  & -25.78 & 32.85 & 7.58  & -25.27 \\
DELLA           & 32.36 & 8.01  & -24.35 & 32.75 & 7.59  & -25.16 \\
Breadcrumbs     & 35.98 & 24.64 & -11.34 & 34.34 & 23.99 & -10.35 \\
Task Arithmetic & 34.98 & 26.60 & -8.38  & 33.88 & 23.60 & -10.28 \\
EMR-Merging     & 35.19 & 26.13 & -9.06  & 33.65 & 23.19 & -10.46 \\
LoRaHub         & 34.70 & 33.44 & -1.26  & 34.80 & 33.11 & -1.69  \\
\textbf{\texttt{Evo-Merging}} & \textbf{39.09} & \textbf{43.59} & \textbf{+4.50} & \textbf{38.03} & \textbf{42.20} & \textbf{+4.17} \\
\bottomrule
\end{tabular}
\vspace{-2mm}
\caption{Average performance on 8-task and 13-task merging. The "Drop" column shows the performance change from 8 to 13 tasks.}
\label{tab:drop_performance_direct}
\end{table}


\paragraph{Scalability with an Increasing Number of Models.}

\begin{figure}[t]
    \centering
    \includegraphics[width=1\linewidth]{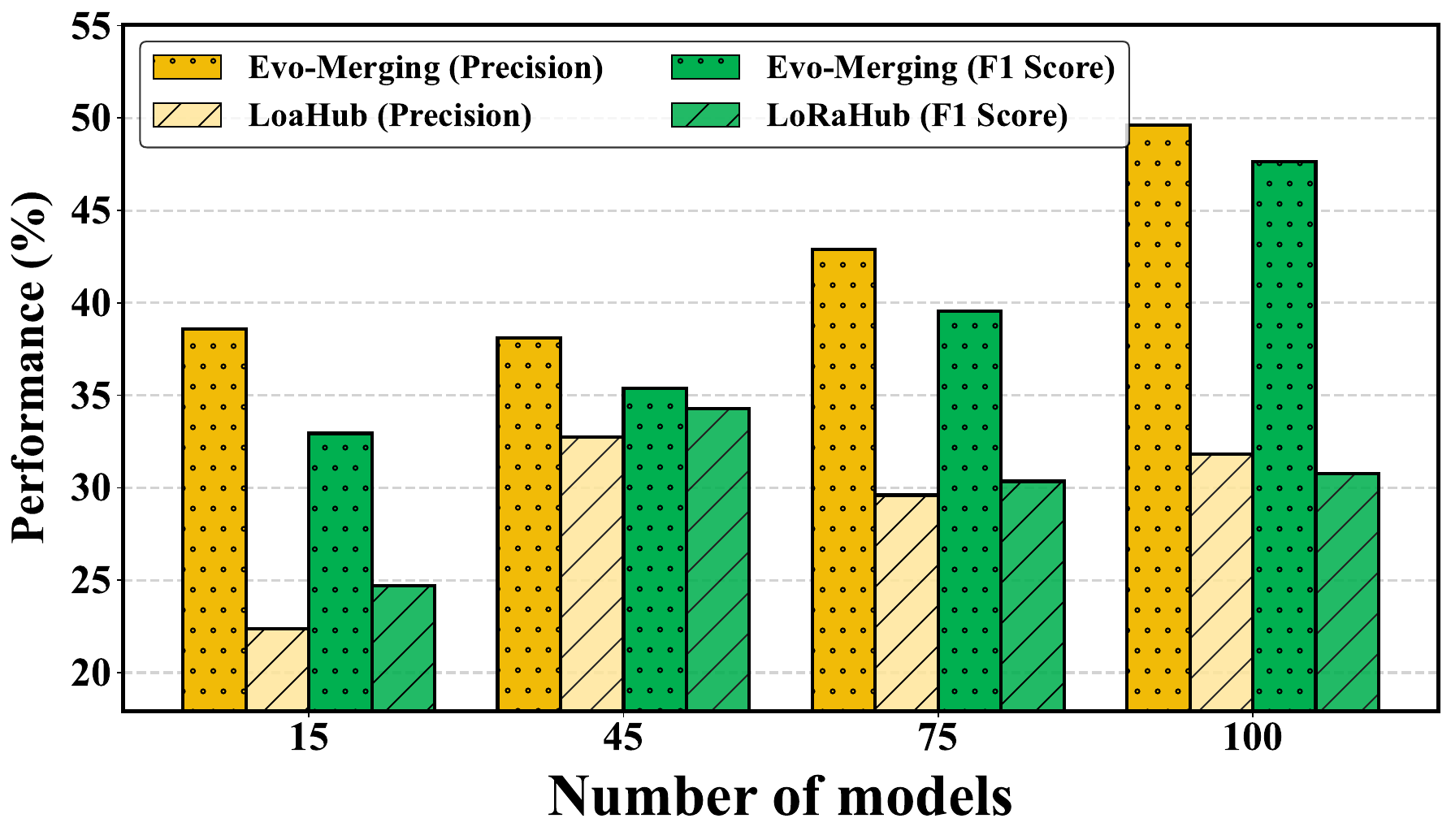}
    \vspace{-7mm}
     \caption{Influence of number of models.}
    \label{fig:Adapter_Ablation}
    \vspace{-2mm}
\end{figure}

To assess scalability, we compare \texttt{Evo-Merging} against the strong baseline LoRaHub by fusing an incrementally larger set of randomly sampled models (from 15 to 100). As illustrated in Figure~\ref{fig:Adapter_Ablation}, \texttt{Evo-Merging}'s performance scales positively with the number of models. This indicates it effectively leverages beneficial knowledge from the expanding pool. Conversely, LoRaHub's performance degrades, revealing an inability to resolve the increased signal conflicts inherent in larger-scale fusion. This comparison confirms \texttt{Evo-Merging}'s superior performance and robustness in complex, many-model scenarios.

\paragraph{Effect of Validation Data Sample Size.}

\begin{figure}[t]
    \centering
    \includegraphics[width=0.6\linewidth]{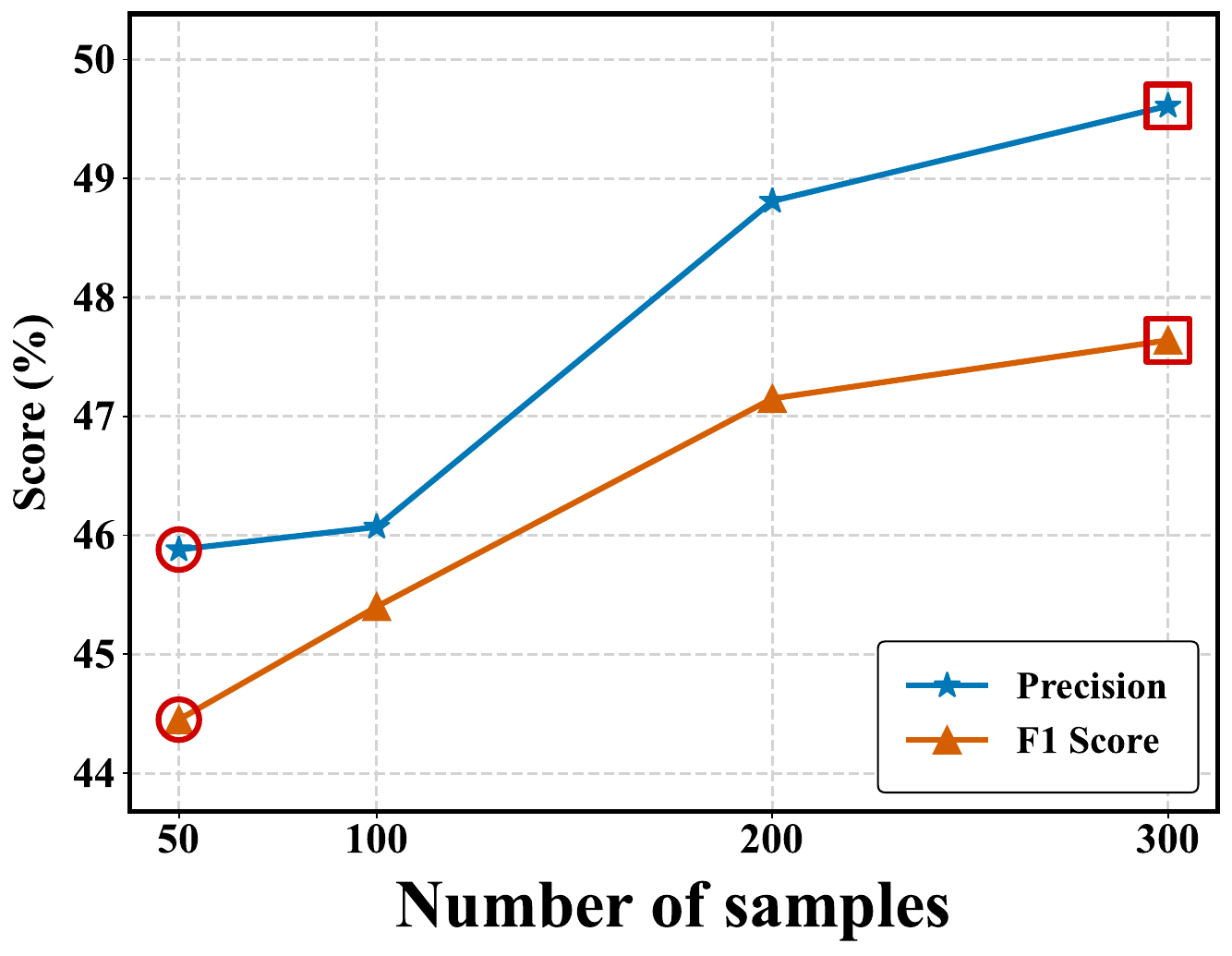}
    \vspace{-1mm}
    \caption{Influence of sample size.}
    \label{fig:Performance_vs._Sample_Size}
    \vspace{-4mm}
\end{figure}

We evaluate the influence of sample number on the NER\_FindVehicle~\cite{guan2022} dataset (Figure~\ref{fig:Performance_vs._Sample_Size}). We observe that \texttt{Evo-Merging} achieves a 44.45\% F1 with only 50 samples, outperforming most baselines that rely on significantly more data. Performance improves steadily with larger sample sizes and the model generally achieves relatively good performance when the sample size reaches around 300, highlighting the method's high data efficiency and strong effectiveness.

\begin{figure}[t]
    \centering
    \includegraphics[width=1\linewidth]{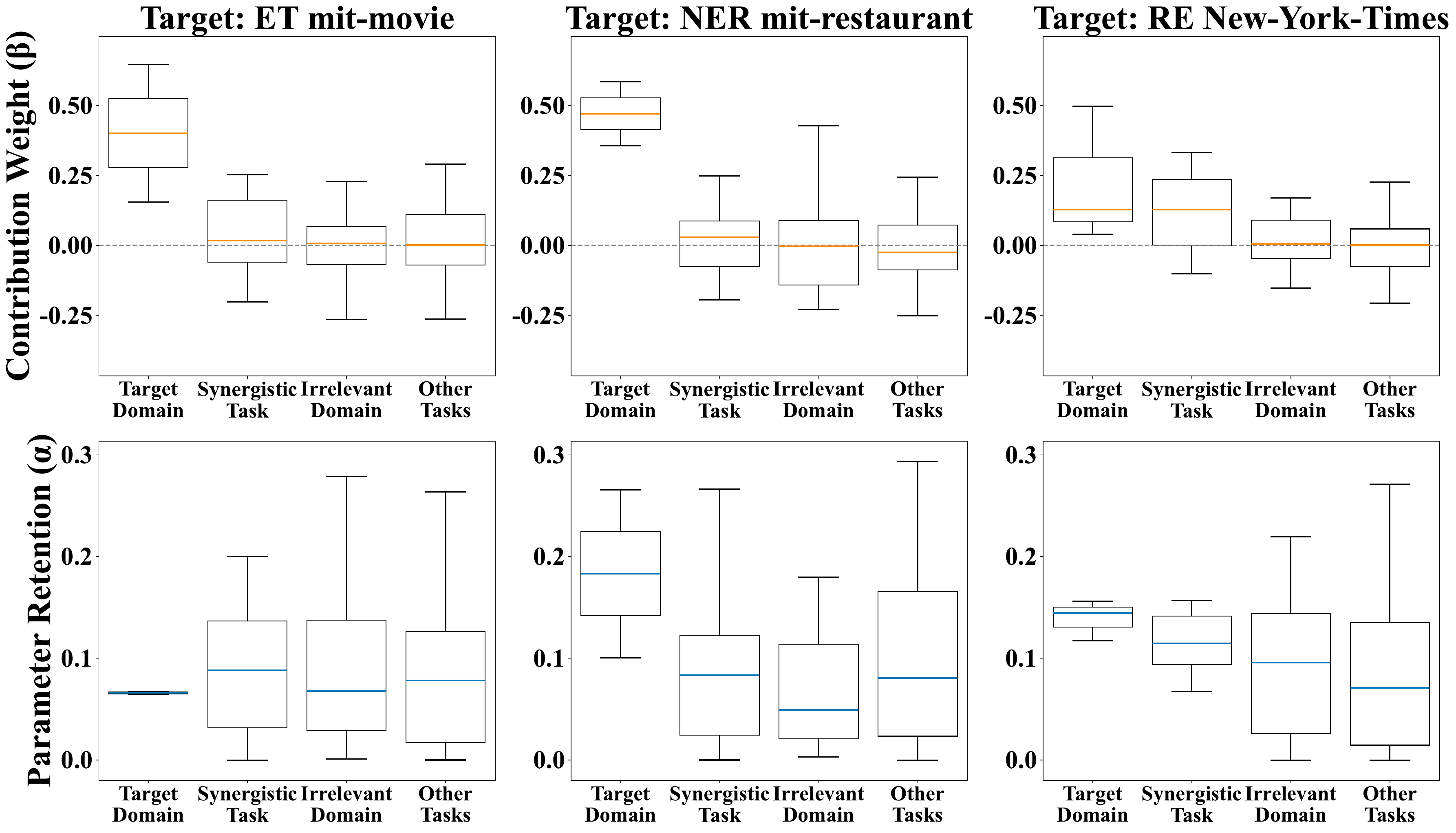}
    \vspace{-5mm}
    \caption{Analysis of learned parameters ($\alpha$, $\beta$) across three tasks. The top row shows contribution weights ($\beta$) and the bottom shows retention ratios ($\alpha$).}
    \label{fig:fusion_solution_analysis}
    \vspace{-1mm}
\end{figure}

\begin{figure}[t!]
    \centering
    \includegraphics[width=1\linewidth]{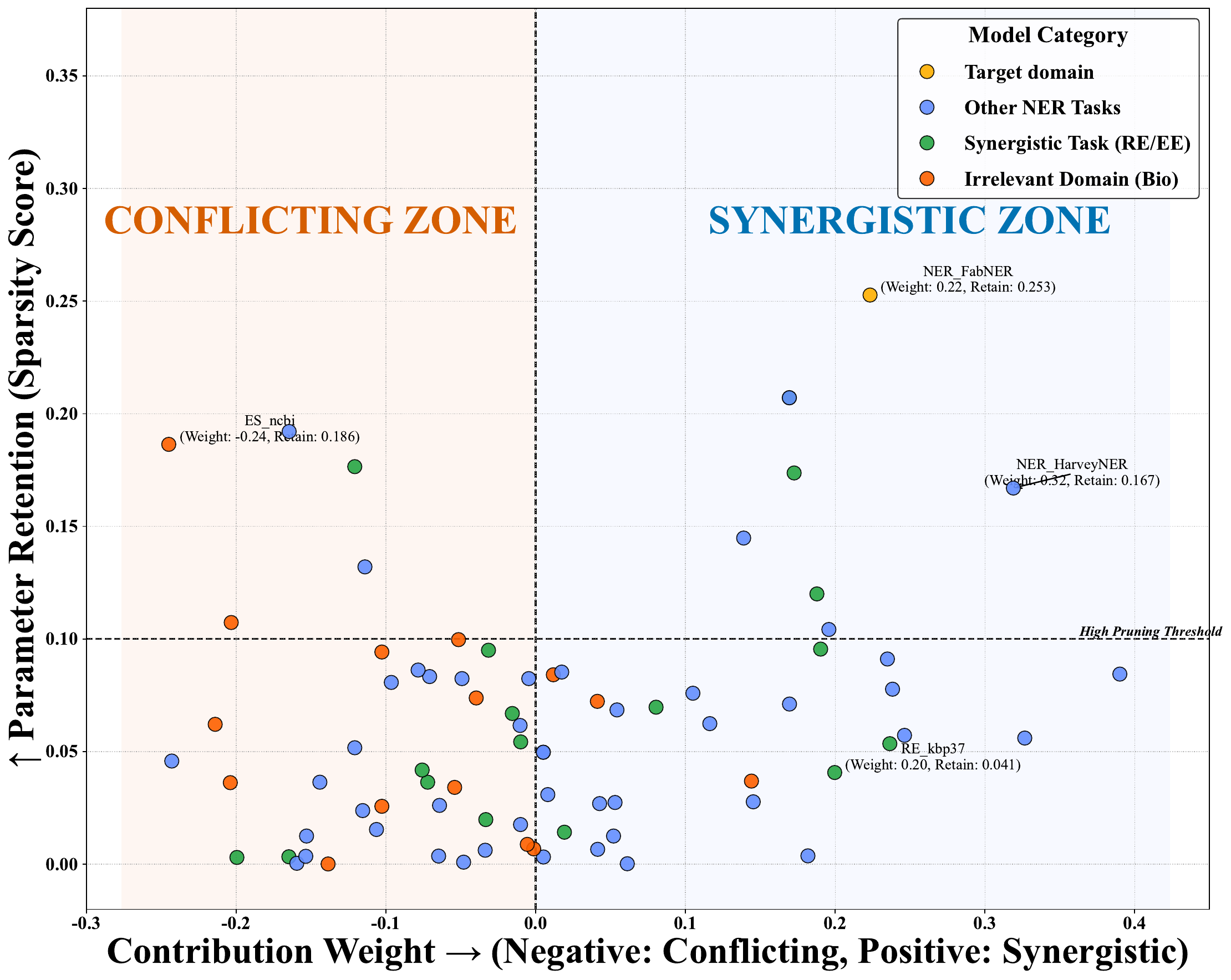}
    \vspace{-5mm}
    \caption{Fusion results over NER\_FindVehicle. Each dot is a source model, positioned by its contribution weights ($\beta$) and retention ratios ($\alpha$).}
    \label{fig:fusion_analysis}
    \vspace{-4mm}
\end{figure}

\paragraph{Analysis of the Fusion Solution}
\label{sec: Effect of Sparsity-Based Denoising and Sign-Aware Scaling.}
We give an analysis to evaluate the effectiveness of our framework (Figures \ref{fig:fusion_solution_analysis} and \ref{fig:fusion_analysis}). 
First, sparsity-based denoising removes redundant parameters from all source models. Models from \textbf{Target} and \textbf{Synergistic Tasks} retain most of their parameters ($\alpha$), while those from \textbf{Irrelevant Domains} are heavily pruned.  
In the second stage, sign-aware scaling assigns contribution weights ($\beta$) intelligently. It amplifies useful models with large positive weights, while driving the weights of irrelevant models toward zero or even negative values. 
This enables the framework to actively suppress conflicting knowledge by repurposing irrelevant models as corrective signals.  



\section{Conclusions and Future Work}

In this paper, we introduce Black-box Model Merging (\texttt{BMM}), a practical paradigm for fusing models accessible only through APIs. We propose \texttt{Evo-Merging}, a novel evolutionary framework that operates without parameter access, using a two-stage process of sparsity-based denoising and sign-aware scaling to intelligently filter noise and resolve conflicts among massive model repositories. Extensive experiments show that \texttt{Evo-Merging} sets a new state-of-the-art, significantly outperforming baselines in both out-of-domain and in-domain scenarios. Our method demonstrates exceptional robustness and scalability, effectively merging over 100 models where prior approaches degrade. 
In the future, we would like to generalize \texttt{Evo-Merging} to merge black-box multi-modal models by designing evolutionary search strategies over cross-modal representations.

\newpage
\clearpage
\bibliography{aaai2026}

\begin{thebibliography}{35}
\providecommand{\natexlab}[1]{#1}

\bibitem[{Akiba et~al.(2025)Akiba, Shing, Tang, Sun, and Ha}]{akiba2025evolutionary}
Akiba, T.; Shing, M.; Tang, Y.; Sun, Q.; and Ha, D. 2025.
\newblock Evolutionary optimization of model merging recipes.
\newblock \emph{Nature Machine Intelligence}, 1--10.

\bibitem[{Benedek and Wolf(2024)}]{benedek2024priLoRaprunedrankincreasinglowrank}
Benedek, N.; and Wolf, L. 2024.
\newblock PRILoRA: Pruned and Rank-Increasing Low-Rank Adaptation.
\newblock arXiv:2401.11316.

\bibitem[{Davari and Belilovsky(2024)}]{davari2024modelbreadcrumbsscalingmultitask}
Davari, M.; and Belilovsky, E. 2024.
\newblock Model Breadcrumbs: Scaling Multi-Task Model Merging with Sparse Masks.
\newblock arXiv:2312.06795.

\bibitem[{Deep, Bhardwaj, and Poria(2024)}]{deep2024dellamergingreducinginterferencemodel}
Deep, P.~T.; Bhardwaj, R.; and Poria, S. 2024.
\newblock DELLA-Merging: Reducing Interference in Model Merging through Magnitude-Based Sampling.
\newblock arXiv:2406.11617.

\bibitem[{Guan(2022)}]{guan2022}
Guan, R. 2022.
\newblock Findvehicle and vehiclefinder: A ner dataset for a text-image cross-modal vehicle retrieval system.
\newblock Dataset.

\bibitem[{Hansen and Ostermeier(1996)}]{hansen1996adapting}
Hansen, N.; and Ostermeier, A. 1996.
\newblock Adapting arbitrary normal mutation distributions in evolution strategies: The covariance matrix adaptation.
\newblock In \emph{Proceedings of IEEE International Conference on Evolutionary Computation (ICEC)}, 312--317. IEEE.

\bibitem[{Hu et~al.(2021)Hu, Shen, Wallis, Allen-Zhu, Li, Wang, Wang, and Chen}]{hu2021LoRa}
Hu, E.~J.; Shen, Y.; Wallis, P.; Allen-Zhu, Z.; Li, Y.; Wang, S.; Wang, L.; and Chen, W. 2021.
\newblock Lora: Low-rank adaptation of large language models.
\newblock In \emph{International Conference on Learning Representations}.

\bibitem[{Huang et~al.(2024{\natexlab{a}})Huang, Liu, Lin, Pang, Du, and Lin}]{huang2024LoRaHubefficientcrosstaskgeneralization}
Huang, C.; Liu, Q.; Lin, B.~Y.; Pang, T.; Du, C.; and Lin, M. 2024{\natexlab{a}}.
\newblock LoraHub: Efficient Cross-Task Generalization via Dynamic LoRA Composition.
\newblock arXiv:2307.13269.

\bibitem[{Huang et~al.(2024{\natexlab{b}})Huang, Ye, Chen, He, Yue, and Ouyang}]{huang2024emr}
Huang, C.; Ye, P.; Chen, T.; He, T.; Yue, X.; and Ouyang, W. 2024{\natexlab{b}}.
\newblock Emr-merging: Tuning-free high-performance model merging.
\newblock \emph{Advances in Neural Information Processing Systems}, 37: 122741--122769.

\bibitem[{Ilharco et~al.(2023{\natexlab{a}})Ilharco, Ribeiro, Wortsman, Gururangan, Schmidt, and Hajishirzi}]{ilharco2023editing}
Ilharco, G.; Ribeiro, M.~T.; Wortsman, M.; Gururangan, S.; Schmidt, L.; and Hajishirzi, H. 2023{\natexlab{a}}.
\newblock Editing Models with Task Arithmetic.
\newblock In \emph{The Eleventh International Conference on Learning Representations}.

\bibitem[{Ilharco et~al.(2023{\natexlab{b}})Ilharco, Ribeiro, Wortsman, Gururangan, Schmidt, Hajishirzi, and Farhadi}]{ilharco2022editing}
Ilharco, G.; Ribeiro, M.~T.; Wortsman, M.; Gururangan, S.; Schmidt, L.; Hajishirzi, H.; and Farhadi, A. 2023{\natexlab{b}}.
\newblock Editing models with task arithmetic.
\newblock In \emph{The Eleventh International Conference on Learning Representations}.

\bibitem[{Kumar and Starly(2021)}]{kumar2021fabner}
Kumar, A.; and Starly, B. 2021.
\newblock "fabner": information extraction from manufacturing process science domain literature using named entity recognition.
\newblock \emph{Journal of Intelligent Manufacturing}, 33: 2393--2407.

\bibitem[{Liu et~al.(2022)Liu, Tam, Muqeeth, Mohta, Huang, Bansal, and Raffel}]{liu2022few}
Liu, H.; Tam, D.; Muqeeth, M.; Mohta, J.; Huang, T.; Bansal, M.; and Raffel, C. 2022.
\newblock Few-shot parameter-efficient fine-tuning is better and cheaper than in-context learning.
\newblock In \emph{Advances in Neural Information Processing Systems}, volume~35, 1950--1965.

\bibitem[{Liu et~al.(2019)Liu, Meng, Zhang, Xu, Chen, and Zhou}]{liu2019}
Liu, Y.; Meng, F.; Zhang, J.; Xu, J.; Chen, Y.; and Zhou, J. 2019.
\newblock {GCDT:} {A} Global Context Enhanced Deep Transition Architecture for Sequence Labeling.
\newblock \emph{CoRR}, abs/1906.02437.

\bibitem[{Liu et~al.(2020)Liu, Xu, Yu, Dai, Ji, Cahyawijaya, Madotto, and Fung}]{liu2020crossner}
Liu, Z.; Xu, Y.; Yu, T.; Dai, W.; Ji, Z.; Cahyawijaya, S.; Madotto, A.; and Fung, P. 2020.
\newblock Crossner: Evaluating cross-domain named entity recognition.
\newblock \emph{CoRR}, abs/2005.02455.

\bibitem[{Lu et~al.(2024)Lu, Fan, Wei, Qu, Chen, and Cheng}]{lu2024twin}
Lu, Z.; Fan, C.; Wei, W.; Qu, X.; Chen, D.; and Cheng, Y. 2024.
\newblock Twin-merging: Dynamic integration of modular expertise in model merging.
\newblock \emph{Advances in Neural Information Processing Systems}, 37: 78905--78935.

\bibitem[{Matena and Raffel(2022)}]{matena2022merging}
Matena, M.~S.; and Raffel, C.~A. 2022.
\newblock Merging models with fisher-weighted averaging.
\newblock \emph{Advances in Neural Information Processing Systems}, 35: 17703--17716.

\bibitem[{Miyano and Arase(2025)}]{miyano2025adaptiveLoRamergeparameter}
Miyano, R.; and Arase, Y. 2025.
\newblock Adaptive LoRA Merge with Parameter Pruning for Low-Resource Generation.
\newblock \emph{arXiv preprint arXiv:2505.24174}.

\bibitem[{Pan et~al.(2017)Pan, Zhang, May, Nothman, Knight, and Ji}]{pan2017cross}
Pan, X.; Zhang, B.; May, J.; Nothman, J.; Knight, K.; and Ji, H. 2017.
\newblock Cross-lingual name tagging and linking for 282 languages.
\newblock In \emph{Proceedings of the 55th Annual Meeting of the Association for Computational Linguistics (Volume 1: Long Papers)}, 1946--1958.

\bibitem[{Riedel, Yao, and McCallum(2010)}]{riedel2010}
Riedel, S.; Yao, L.; and McCallum, A. 2010.
\newblock Modeling relations and their mentions without labeled text.
\newblock In \emph{Proceedings of the European Conference on Machine Learning and Principles and Practice of Knowledge Discovery in Databases (ECML PKDD)}, 148--163.

\bibitem[{Roth and tau Yih(2004)}]{roth2004}
Roth, D.; and tau Yih, W. 2004.
\newblock A linear programming formulation for global inference in natural language tasks.
\newblock In \emph{Proceedings of the Eighth Conference on Computational Natural Language Learning (CoNLL-2004)}, 1--8.

\bibitem[{Si, Yang, and Shen(2024)}]{si2024parameterefficientfinetuningstanding}
Si, C.; Yang, X.; and Shen, W. 2024.
\newblock See Further for Parameter Efficient Fine-tuning by Standing on the Shoulders of Decomposition.
\newblock arXiv:2407.05417.

\bibitem[{Stoica et~al.(2024)Stoica, Ramesh, Ecsedi, Choshen, and Hoffman}]{stoica2024modelmergingsvdtie}
Stoica, G.; Ramesh, P.; Ecsedi, B.; Choshen, L.; and Hoffman, J. 2024.
\newblock Model merging with svd to tie the knots.
\newblock \emph{arXiv preprint arXiv:2410.19735}.

\bibitem[{Sun et~al.(2022)Sun, Shao, Qian, Huang, and Qiu}]{sun2022black}
Sun, T.; Shao, Y.; Qian, H.; Huang, X.; and Qiu, X. 2022.
\newblock Black-box tuning for language-model-as-a-service.
\newblock In \emph{International Conference on Machine Learning}, 20841--20855. PMLR.

\bibitem[{Tedeschi and Navigli(2022)}]{tedeschi-navigli-2022-multinerd}
Tedeschi, S.; and Navigli, R. 2022.
\newblock {M}ulti{NERD}: A Multilingual, Multi-Genre and Fine-Grained Dataset for Named Entity Recognition (and Disambiguation).
\newblock In \emph{Findings of the Association for Computational Linguistics: NAACL 2022}, 801--812. Seattle, United States: Association for Computational Linguistics.

\bibitem[{{The Llama 3.1 Team}(2024)}]{meta2024llama3.1}
{The Llama 3.1 Team}. 2024.
\newblock {The Llama 3.1 Series of Models}.
\newblock arXiv:2407.18342.

\bibitem[{Ushio et~al.(2022)Ushio, Neves, Silva, Barbieri, and Camacho-Collados}]{ushio2022named}
Ushio, A.; Neves, L.; Silva, V.; Barbieri, F.; and Camacho-Collados, J. 2022.
\newblock {Named Entity Recognition in Twitter: A Dataset and Analysis on Short-Term Temporal Shifts}.
\newblock In \emph{Proceedings of the 2nd Conference of the Asia-Pacific Chapter of the Association for Computational Linguistics and the 12th International Joint Conference on Natural Language Processing}, 309--319.

\bibitem[{Walker et~al.(2006)Walker, Strassel, Medero, and Maeda}]{walker2006ace}
Walker, C.; Strassel, S.; Medero, J.; and Maeda, K. 2006.
\newblock {ACE 2005 Multilingual Training Corpus}.
\newblock In \emph{LDC Catalog}. Linguistic Data Consortium.

\bibitem[{Wang et~al.(2024)Wang, Ping, Wang, Han, Chen, Liu, and Sun}]{wang2024LoRaflowdynamicLoRafusion}
Wang, H.; Ping, B.; Wang, S.; Han, X.; Chen, Y.; Liu, Z.; and Sun, M. 2024.
\newblock Lora-flow: Dynamic lora fusion for large language models in generative tasks.
\newblock \emph{arXiv preprint arXiv:2402.11455}.

\bibitem[{Wang et~al.(2023)Wang, Zhou, Zu, Xia, Chen, Zhang, Zheng, Ye, Zhang, Gui, Kang, Yang, Li, and Du}]{wang2023instructuiemultitaskinstructiontuning}
Wang, X.; Zhou, W.; Zu, C.; Xia, H.; Chen, T.; Zhang, Y.; Zheng, R.; Ye, J.; Zhang, Q.; Gui, T.; Kang, J.; Yang, J.; Li, S.; and Du, C. 2023.
\newblock InstructUIE: Multi-task Instruction Tuning for Unified Information Extraction.
\newblock arXiv:2304.08085.

\bibitem[{Yadav et~al.(2023)Yadav, Tam, Choshen, Raffel, and Bansal}]{yadav2023ties}
Yadav, P.; Tam, D.; Choshen, L.; Raffel, C.~A.; and Bansal, M. 2023.
\newblock Ties-merging: Resolving interference when merging models.
\newblock \emph{Advances in Neural Information Processing Systems}, 36: 7093--7115.

\bibitem[{Yang et~al.(2024)Yang, Shen, Guo, Wang, Cao, Zhang, and Tao}]{yang2024model}
Yang, E.; Shen, L.; Guo, G.; Wang, X.; Cao, X.; Zhang, J.; and Tao, D. 2024.
\newblock Model merging in llms, mllms, and beyond: Methods, theories, applications and opportunities.
\newblock \emph{arXiv preprint arXiv:2408.07666}.

\bibitem[{Yu et~al.(2024{\natexlab{a}})Yu, Yu, Yu, Huang, and Li}]{yu2024language}
Yu, L.; Yu, B.; Yu, H.; Huang, F.; and Li, Y. 2024{\natexlab{a}}.
\newblock Language models are super mario: Absorbing abilities from homologous models as a free lunch.
\newblock In \emph{Forty-first International Conference on Machine Learning}.

\bibitem[{Yu et~al.(2024{\natexlab{b}})Yu, Yu, Yu, Huang, and Li}]{yu2024languagemodelssupermario}
Yu, L.; Yu, B.; Yu, H.; Huang, F.; and Li, Y. 2024{\natexlab{b}}.
\newblock Language Models are Super Mario: Absorbing Abilities from Homologous Models as a Free Lunch.
\newblock arXiv:2311.03099.

\bibitem[{Zhao et~al.(2024)Zhao, Shen, Zhu, Li, Su, Wang, Kuang, and Wu}]{zhao2024mergingLoRaslikeplaying}
Zhao, Z.; Shen, T.; Zhu, D.; Li, Z.; Su, J.; Wang, X.; Kuang, K.; and Wu, F. 2024.
\newblock Merging LoRAs like Playing LEGO: Pushing the Modularity of LoRA to Extremes Through Rank-Wise Clustering.
\newblock arXiv:2409.16167.

\end{thebibliography}

\newpage
\clearpage
\setcounter{secnumdepth}{2}

\appendix 

\section{Evo-Merging Implementation Details}
\label{sec:Evo-Merging Implement Detail}

\paragraph{Experimental Environment.}
 For the model merging stage, we utilized an Intel(R) Xeon(R) Platinum 8352Y CPU @ 2.20GHz. For the inference stage, the resulting models were deployed on  GPUs: an NVIDIA A800 (80 GB VRAM).

\paragraph{Core Optimization Parameters.}
Our \texttt{Evo-Merging} framework is built on the Covariance Matrix Adaptation Evolution Strategy (CMA-ES)~\cite{hansen1996adapting}, for which we used the \texttt{cma} Python package. Across all experiments, we used a fixed LoRa rank of 8, a CMA-ES population size of 20, a sigma of 0.05, a bound of 1.5, and an L1 regularization coefficient ($\lambda$) of 0.05.

\paragraph{Training and Inference details.}
To construct our repository of over 100 specialized models, we performed fine-tuning on the \textbf{Llama 3.1 8B Instruct} model for each downstream task. We applied the \textbf{AdamW optimizer} throughout all training processes. The training was configured with a batch size of 128, and a maximum sequence length (\texttt{cutoff\_len}) of 2048 tokens. All models were trained for 20 epochs (\texttt{num\_train\_epochs}) with an initial learning rate of 1e-4. We employed a cosine learning rate scheduler (\texttt{lr\_scheduler\_type}) with a warm-up ratio (\texttt{warmup\_ratio}) of 0.1, where the learning rate was gradually increased during the initial 10\% of training steps before following the cosine decay.

For the generation process during inference, we utilized a nucleus sampling strategy with a \texttt{temperature} of 0.95, \texttt{top\_p} of 0.7, and \texttt{top\_k} of 50. The maximum number of newly generated tokens (\texttt{max\_new\_tokens}) was set to 1024.

\paragraph{Experimental Setups and Checkpointing.}
The specific configuration for our two primary experimental settings is summarized in Table~\ref{tab:exp_setups}. We tailored the number of iterations to the complexity of each task. To mitigate overfitting, the loss was accumulated over the entire validation set for each fitness evaluation, and we employed a greedy checkpointing strategy, saving the best model found during the search. The final models reported in this paper correspond to the iteration number listed in the table.
\begin{table}[H]
\centering
\scriptsize
\setlength{\tabcolsep}{0.5mm}
\begin{tabular}{lcc}
\toprule
\textbf{Parameter} & \textbf{OOD Setting} & \textbf{ID Setting} \\ 
\midrule
Model Pool Size & 100+ & N (e.g., 8) \\ 
Validation Set Size & 300 samples (total) & 100 samples per task \\ 
Stage 1 Iterations & 20 & 20 \\ 
Stage 2 Iterations & 40 & 20 \\ 
\textbf{Final Checkpoint} & \textbf{Iteration 34 of Stage 2 } & \textbf{Iteration  2 and 6 of Stage 2} \\ 
\bottomrule
\end{tabular}
\caption{Experimental configurations for OOD and ID settings.}
\label{tab:exp_setups}
\end{table}
We observed that the extended 40-iteration search in Stage 2 was consistently beneficial for scenarios involving more than 50 adapters, justifying our setup for the large-scale OOD task.

\section{Algorithm flow of Evo-Merging}
\label{alg: Algorithm flow of Evo-Merging}
We summarize the procedure of \texttt{Evo-Merging} in Algorithm \ref{alg:Evo-Merging}.
\begin{algorithm}[H]
\caption{Evo-Merging Procedure}
\label{alg:Evo-Merging}
\begin{algorithmic}[1] 
\REQUIRE Collection of LoRa pairs $\{A_i, B_i\}_{i=1..N}$, validation data $\mathcal{D}_{\text{val}}$, regularization parameters $\lambda_1, \lambda_2$.
\ENSURE Merged LoRa pair $(\mathcal{A}_m, \mathcal{B}_m)$.
\STATE 

\STATE $\triangleright$ \textbf{Step 1: Sparsity-Based Denoising}
\STATE Set uniform weights $w_i \gets 1/N$ for $i=1..N$.
\STATE $B_{\text{pm}} \gets \sum_{i=1}^{N} w_i \cdot B_i$.
\STATE Use CMA-ES to find optimal $\bm{\alpha}^*$:

    \STATE \quad $\bm{\alpha}^* \gets \underset{\bm{\alpha} \in [0,1]^N}{\text{argmin}} \Big( \mathcal{L}_{\text{CE}}(\mathcal{M}(A_{\text{pm}}(\bm{\alpha}), B_{\text{pm}}), \mathcal{D}_{\text{val}})$
    \STATE \quad \qquad\qquad\qquad\quad ${} + \lambda_1 ||\bm{\alpha}||_1 \Big)$
    \STATE \quad where $A_{\text{pm}}(\bm{\alpha}) = \sum_{i=1}^{N} w_i \cdot \mathbf{S}_{\alpha_i}(A_i)$.
\STATE

\STATE $\triangleright$ \textbf{Step 2: Sign-Aware Scaling}
\STATE Create sparsified LoRa-A matrices:
\FOR{$i = 1, \dots, N$}
    \STATE $A'_i \gets \mathbf{S}_{\alpha_i^*}(A_i)$
\ENDFOR
\STATE Use CMA-ES to find optimal $\bm{\beta}^*$:
    \STATE \quad $\bm{\beta}^* \gets \underset{\bm{\beta} \in \mathbb{R}^N}{\text{argmin}} \Big( \mathcal{L}_{\text{CE}}(\mathcal{M}(\bm{\beta}), \mathcal{D}_{\text{val}})$
    \STATE \quad \qquad\qquad\qquad ${} + \lambda_2 ||\bm{\beta}||_1 \Big)$
    \STATE \quad where $\mathcal{M}(\bm{\beta})$ uses $\mathcal{A}_m = \sum \beta_i A'_i, \mathcal{B}_m = \sum \beta_i B_i$.
\STATE

\STATE $\triangleright$ \textbf{Step 3: Construct Final Merged Model}
\STATE $\mathcal{A}_m \gets \sum_{i=1}^{N} \beta_i^* \cdot A'_i$
\STATE $\mathcal{B}_m \gets \sum_{i=1}^{N} \beta_i^* \cdot B_i$
\end{algorithmic}
\end{algorithm}

\section{Details of Baselines}
\label{sec: BaseLine Details}
This section provides a detailed description of the baseline methods used for comparison in our experiments. Each method is summarized by its core concept and a simplified mathematical representation.

\begin{description}
    \item[Individual Model] Each task utilizes an independently fine-tuned model, serving as an upper-bound for single-task performance. The model for a specific task $t$ is denoted as $\theta_t$.

    \item[Task Arithmetic \cite{ilharco2022editing}] Merges models by arithmetically combining their “task vectors” ($\tau_t$), which represent the difference between fine-tuned ($\theta_t$) and pre-trained ($\theta_{\text{pre}}$) weights.
    \[
        \theta_m = \theta_{\text{pre}} + \lambda \sum_t \tau_t
    \]

    \item[TIES-Merging \cite{yadav2023ties}] Resolves interference by a three-step process: \textbf{T}rimming low-magnitude parameters, \textbf{E}lecting a dominant sign, and \textbf{S}eparately merging parameter groups that agree on the sign.
    \[
        \tau_m = \text{TIES}(\{\tau_t\})
    \]

    \item[DARE (Drop and Rescale) \cite{yu2024languagemodelssupermario}] Sparsifies task vectors by randomly \textbf{D}ropping a high proportion ($p$) of delta parameters ($\delta$) and \textbf{RE}scaling the survivors by $1/(1-p)$ to preserve the original model's embeddings.
    \[
        \delta' = \frac{\delta \odot (1-m)}{1-p}, \quad \text{where } m \sim \text{Bernoulli}(p)
    \]

    \item[DELLA-Merging \cite{deep2024dellamergingreducinginterferencemodel}] Employs a novel pruning technique, MAGPRUNE, which samples delta parameters for dropping based on their magnitude, followed by sign election and fusion.
    \[
        p(\text{drop}_i) \propto 1/|\delta_i|
    \]

    \item[EMR-Merging \cite{huang2024emr}] Creates a unified model by \textbf{E}lecting a shared parameter base, then generates lightweight task-specific \textbf{M}asks ($M_t$) and \textbf{R}escalers ($\lambda_t$) to align the direction and magnitude for each task.
    \[
        \theta'_t = \lambda_t \cdot (M_t \odot \theta_{\text{unified}})
    \]

    \item[Breadcrumbs \cite{davari2024modelbreadcrumbsscalingmultitask}] Constructs multi-task models by creating a sparse 'breadcrumb' mask ($m_t$) that filters out both low-magnitude and high-magnitude outlier weights from task vectors on a layer-wise basis.
    \[
        \theta_m = \theta_{\text{pre}} + \alpha \sum_t (m_t \odot \tau_t)
    \]

    \item[LoRaHub ~\cite{huang2024LoRaHubefficientcrosstaskgeneralization}] Achieves cross-task generalization by dynamically composing a set of pre-trained LoRA modules ($m_i$) for a new task using a gradient-free optimization to find the optimal combination weights ($w_i$).
    \[
        m_{\text{composed}} = \sum_i w_i m_i
    \]

    \item[LoRa (Low-Rank Adaptation) \cite{hu2021LoRa}] Adapts pre-trained models by freezing the original weights ($W_0$) and injecting trainable low-rank decomposition matrices ($B$ and $A$) into each layer.
    \[
        W \leftarrow W_0 + BA
    \]

    \item[(IA)\textsuperscript{3} \cite{liu2022few}] Injects and learns minimal scaling vectors ($l$) that modulate internal activations ($h$), allowing for efficient adaptation with extremely few trainable parameters.
    \[
        h' = l \odot h
    \]

    \item[Best Adapter] A non-compositional baseline that selects the single best-performing source adapter ($m^*$) on the target task's validation set ($D_{\text{val}}$).
    \[
        m^* = \arg\min_{m_i} \mathcal{L}(m_i, D_{\text{val}})
    \]
\end{description}

\section{Dataset Details and Metric}
\label{sec: Dataset Details and Evalution}
\subsection{Dataset}
All experiments in this paper are conducted using the \textbf{InstructUIE} benchmark~\cite{wang2023instructuiemultitaskinstructiontuning}, a large-scale, unified framework for Information Extraction (IE). The core innovation of InstructUIE is its reformulation of diverse IE tasks into a single, generative text-to-text format. This unification is the key technical enabler that allows us to represent specialist models, each trained on fundamentally different extraction objectives, within a comparable parameter space, making them viable candidates for merging.

Our experimental methodology hinges on the creation of a large and deeply heterogeneous \textbf{source model pool}. We fine-tuned over 100 specialist \texttt{Llama 3.1 Instruct} models, where each model was trained to master one of the specific, granular IE tasks defined by InstructUIE. To provide the reader with a clear understanding of what exactly is being merged in our framework, we detail the hierarchy of these tasks below.

\paragraph{The Hierarchy of Information Extraction Tasks}
The tasks within InstructUIE represent different layers of semantic understanding, from identifying simple entities to deconstructing complex events. Our model pool contains specialists for each of the following task categories:

\begin{itemize}
    \item \textbf{Layer 1: Named Entity Recognition (NER) — The Foundational Layer.} This is the most fundamental IE task, focused on locating and classifying atomic pieces of information—the "who," "what," and "where" in a text.
    \begin{itemize}
        \item \textbf{Primary Task: Named Entity Recognition (NER).} This is the complete task: to both identify an entity's text span and assign it a type (e.g., identifying "Apple Inc." and labeling it as `Organization`).
        \item \textbf{Auxiliary Task: Entity Span Extraction (ES).} A simpler, decomposed version of NER that only focuses on identifying the boundaries of an entity (e.g., finding "Apple Inc.") without knowing its type.
        \item \textbf{Auxiliary Task: Entity Typing (ET).} The other half of the NER problem: given a specific text span like "Apple Inc.", the task is to assign the correct label (`Organization`).
    \end{itemize}

    \item \textbf{Layer 2: Relation Extraction (RE) — Connecting the Dots.} Building upon the foundation of entities, RE focuses on identifying the semantic relationships that connect them.
    \begin{itemize}
        \item \textbf{Primary Task: Relation Extraction (RE).} The complete task of identifying a full semantic triplet, including two entities and the relation between them (e.g., `(Steve Jobs, founded, Apple Inc.)`).
        \item \textbf{Auxiliary Task: Entity Pair Identification (EP).} A simpler task that only identifies that a pair of entities is related, without specifying the nature of the relationship (e.g., identifying that `(Steve Jobs, Apple Inc.)` is a related pair).
        \item \textbf{Auxiliary Task: Entity Pair Relationship Identification (EPR).} The classification component: given a pair of entities `(Steve Jobs, Apple Inc.)`, the task is to classify the relationship as `founded`.
    \end{itemize}

    \item \textbf{Layer 3: Event Extraction (EE) — Deconstructing Structured Scenarios.} This is the most complex layer, focused on extracting information about dynamic occurrences or events, including who did what to whom, when, and where.
    \begin{itemize}
        \item \textbf{Primary Task: Event Extraction (EE).} The complete task of identifying an event, its type, its trigger word, and all its associated arguments with their specific roles (e.g., for a `Purchase` event, identifying the `Buyer`, `Seller`, and `Item`).
        \item \textbf{Auxiliary Task: Event Trigger Identification (EET).} The first step in EE: identifying the key word or phrase that triggers the event (e.g., the word "bought") and classifying the event's type (`Purchase`).
        \item \textbf{Auxiliary Task: Event Argument Extraction (EEA).} The second step: given an event trigger and its type, the task is to find all the entities participating in that event and assign them their correct roles (e.g., identifying `Google` as the `Buyer` and `YouTube` as the `Item`).
    \end{itemize}
\end{itemize}
Crucially, our source pool of 100+ models is populated with specialists from \textbf{all nine of these distinct tasks}. The core challenge for our \texttt{Evo-Merging} framework, particularly in the Out-of-Domain setting, is therefore to intelligently fuse knowledge from this highly diverse collection. It must learn how to leverage the foundational knowledge of an \textbf{ES} model, the relational logic of an \textbf{EPR} model, and the structural understanding of an \textbf{EEA} model to generalize effectively to a new, unseen task.

\paragraph{Evaluation Datasets}
Our evaluations are conducted in two settings. The \textbf{Cross-domain} evaluation assesses multi-task capability by merging and testing on eight NER datasets: ACE\_2005~\citep{walker2006ace}, CrossNER\_AI~\cite{liu2020crossner}, CrossNER\_Lit~\cite{liu2020crossner}, FabNER~\cite{kumar2021fabner}, Mit-Movie, MultiNERD~\cite{tedeschi-navigli-2022-multinerd}, TweetNER~\cite{ushio2022named}, and WikiANN~\cite{pan2017cross}. The \textbf{out-of-domain} evaluation, our core test for zero-shot generalization, uses six datasets that are entirely unseen by the source model pool. These include two datasets for NER (\textbf{MIT Restaurant Review}~\cite{liu2019} and \textbf{FindVehicle}~\cite{guan2022}), two for RE (\textbf{NYT}~\cite{riedel2010} and \textbf{CoNLL2004}~\cite{roth2004}), and two for Event Trigger (ET) identification (\textbf{Mit-Movie}~\cite{liu2019} and \textbf{FabNER}~\cite{kumar2021fabner}).

\subsection{Evaluation Metrics}

We evaluate all models using Macro-averaged F1-score and Precision. These metrics are standard for information extraction, effectively handling the typical class imbalance of the task. Precision measures the reliability of LLM's generation.The F1-score provides a single, balanced measure of overall performance. We report the macro-average, which computes the metric per instance before averaging, to ensure every sample contributes equally to the final score.

Our evaluation follows a strict matching criterion, where entities, relations, or triggers are correct only if all their components (e.g., span and type) exactly match the ground truth. To ensure a fair and robust comparison, this process is normalized:
\begin{itemize}
    \item \textbf{Case-Insensitive:} All text is converted to lowercase before comparison, preventing capitalization differences from affecting scores.
    \item \textbf{Order-Insensitive:} The set of extracted items is treated as unordered, meaning the generation order does not influence the result.
\end{itemize}

\section{Pre-selection Strategy for Baselines in OOD Setting}
\label{sec: Pre-selection Strategy for Baselines in OOD Setting}
The large(over 100), noisy model pool in the out-of-domain (OOD) setting poses a significant challenge for most baseline methods, leading to performance degradation and making hyperparameter tuning infeasible.We implemented the following two-stage pre-selection protocol exclusively for these baselines:

\begin{description}
    \item[Adapter Pre-selection:] For each target OOD task, we first identified a promising subset of source adapters (typically less than 10) based on their generalization performance on the same validation set. This step filters for the most relevant models. The resulting curated adapter sets used for each task are detailed in Table~\ref{tab:model_distribution_standardized}.

    \item[Hyperparameter Tuning:] On these manageable subsets, we then performed an extensive grid search to find the optimal hyperparameters for each baseline (e.g., density and $\lambda$). The best-performing configurations, which were used in our main results, are summarized in Table~\ref{sec:appendix_hyperparams}.
\end{description}

In stark contrast, this pre-processing pipeline was not applied to our \texttt{Evo-Merging} framework or LoRaHub. Both were benchmarked directly on the entire pool of 100+ models, rigorously testing their inherent robustness and ability to navigate noisy, large-scale fusion scenarios.

\begin{table*}[t!]
\centering
\setlength{\tabcolsep}{2.5pt} 
\footnotesize 
\begin{tabular}{l|ccc|ccc|ccc|ccc}
\toprule
\multirow{2}{*}{\textbf{Dataset}} & \multicolumn{3}{c|}{\textbf{TIES-Merging}} & \multicolumn{3}{c|}{\textbf{DARE}} & \multicolumn{3}{c|}{\textbf{Della}} & \multicolumn{3}{c}{\textbf{Breadcrumbs}} \\
\cmidrule(lr){2-4} \cmidrule(lr){5-7} \cmidrule(lr){8-10} \cmidrule(l){11-13}
& \textbf{Params ($\lambda, p$)} & \textbf{Prec.} & \textbf{F1} & \textbf{Params ($\lambda, p$)} & \textbf{Prec.} & \textbf{F1} & \textbf{Params ($\lambda, p, \epsilon$)} & \textbf{Prec.} & \textbf{F1} & \textbf{Params ($\lambda, p, \gamma$)} & \textbf{Prec.} & \textbf{F1} \\
\midrule
NER-mitrestaurant & (1.0, 0.9) & 21.40 & 24.22 & (1.0, 0.9) & 10.88 & 14.31 & (0.5, 0.9, 0.1) & 29.69 & 26.06 & (0.5, 0.7, 0.01) & 38.99 & 34.06 \\
NER-FindVehicle   & (1.0, 0.9) & 29.87 & 29.45 & (0.3, 0.9) & 24.13 & 23.69 & (0.7, 0.9, 0.1) & 16.66 & 17.04 & (0.5, 0.9, 0.01) & 47.41 & 38.42 \\
ET-mitmovie       & (0.9, 0.9) & 48.67 & 51.12 & (0.1, 0.5) & 44.57 & 48.60 & (0.7, 0.9, 0.1) & 43.85 & 47.68 & (0.5, 0.7, 0.01) & 49.73 & 51.60 \\
ET-FabNER         & (0.5, 0.9) & 15.91 & 11.42 & (0.1, 0.9) & 15.44 & 10.71 & (1.0, 0.9, 0.1) & 14.41 & 12.51 & (0.5, 0.9, 0.01) & 20.53 & 16.84 \\
RE-NYT            & (1.0, 0.9) & 12.55 & 7.57  & (0.1, 0.9) & 3.69  & 2.53  & (0.5, 0.9, 0.1) & 3.94  & 3.25  & (0.5, 0.9, 0.01) & 34.05 & 28.03 \\
RE-CoNLL04        & (1.0, 0.9) & 20.78 & 19.05 & (0.1, 0.9) & 10.85 & 8.08  & (1.0, 0.9, 0.1) & 19.91 & 17.97 & (0.5, 0.9, 0.01) & 26.30 & 24.56 \\
\bottomrule
\end{tabular}
\caption{Summary of hyperparameter grid search for baseline methods. }
\label{tab:grid_search_summary}
\end{table*}

\begin{table*}[t!]
\centering
\scriptsize 
\setlength{\tabcolsep}{1pt} 
\begin{tabular}{llllll} 
\toprule
\textbf{Target: MIT-Restaurant} & \textbf{Target: FindVehicle} & \textbf{Target: New York Times} & \textbf{Target: CoNLL04} & \textbf{Target: MIT-Movie} & \textbf{Target: FabNER} \\
\midrule
NER CrossNER AI         & NER MIT-Movie SFT       & NER CoNLL 2003     & EPR CoNLL04         & NER MIT-Movie           & ET MultiNERD \\
NER CrossNER Literature & NER CoNLL 2003          & EPR New York Times & EPR SciERC          & NER MIT-Restaurant      & ET CoNLL 2003 \\
NER CrossNER Music      & NER OntoNotes           & RE NYT11           & EPR New York Times  & NER FindVehicle         & ET MIT-Movie \\
NER CrossNER Politics   & NER CrossNER Politics   &                    & RE SciERC           & NER FabNER              & ET CrossNER Science \\
NER CrossNER Science    & NER Broad Tweet Corpus  &                    &                     & NER CrossNER Literature & ET Broad Tweet Corpus \\
NER FabNER              & NER WikiANN EN          &                    &                     & ET CrossNER AI          & ET OntoNotes \\
NER FindVehicle         & NER CrossNER Science    &                    &                     & NER CrossNER AI         & ET WikiANN EN \\
NER MIT-Movie           & NER TweetNER7           &                    &                     & ET CrossNER Music       & ET WikiNeural \\
NER TweetNER7           & NER ACE 2005            &                    &                     & NER CrossNER Music      & ET FindVehicle \\
NER WikiANN EN          & NER MultiNERD           &                    &                     & ET CrossNER Literature  & ET MIT-Restaurant \\
\bottomrule
\end{tabular}
\caption{Standardized Model Distribution Across Datasets. This table shows the top-performing source adapters selected via the pre-selection process for each target OOD task.}
\label{tab:model_distribution_standardized}
\end{table*}

\begin{table*}[t!]
\footnotesize  
\centering
\vspace{-2mm}
\setlength{\tabcolsep}{2.0pt} 
\begin{tabular}{l *{9}{cc}}
\toprule 
\multirow{2}{*}{\textbf{Method}}
& \multicolumn{2}{c}{ACE\_2005} & \multicolumn{2}{c}{CrossNER\_AI} & \multicolumn{2}{c}{CrossNER\_Lit} & \multicolumn{2}{c}{FabNER} & \multicolumn{2}{c}{Mit-Movie} & \multicolumn{2}{c}{MultiNERD} & \multicolumn{2}{c}{TweetNER} & \multicolumn{2}{c}{WikiANN} & \multicolumn{2}{c}{Avg} \\
\cmidrule(lr){2-3} \cmidrule(lr){4-5} \cmidrule(lr){6-7} \cmidrule(lr){8-9} \cmidrule(lr){10-11} \cmidrule(lr){12-13} \cmidrule(lr){14-15} \cmidrule(lr){16-17} \cmidrule(lr){18-19} 
& Prec & F1 & Prec & F1 & Prec & F1 & Prec & F1 & Prec & F1 & Prec & F1 & Prec & F1 & Prec & F1 & Prec & F1 \\
\midrule
TIES & 8.33 & 6.26 & 4.56 & 3.82 & 4.69 & 3.82 & 7.00 & 5.55 & 6.02 & 5.14 & 20.93 & 22.87 & 10.73 & 9.94 & 40.27 & 38.99 & 12.82 & 12.05 \\
DARE & 5.62 & 4.48 & 1.89 & 1.51 & 1.41 & 1.28 & 3.54 & 2.93 & 2.62 & 2.17 & 11.63 & 12.55 & 5.61 & 5.23 & 31.44 & 30.52 & 7.97 & 7.58 \\
DELLA & 5.65 & 4.47 & 1.52 & 1.12 & 1.46 & 1.25 & 3.88 & 3.11 & 2.52 & 2.17 & 12.13 & 13.18 & 5.83 & 5.32 & 31.11 & 30.13 & 8.01 & 7.59 \\
Breadcrumbs & 17.04 & 18.15 & 21.56 & 14.90 & 30.82 & 27.09 & 11.00 & 6.06 & 20.80 & 15.73 & 34.03 & 47.72 & 27.07 & 22.90 & 43.06 & 39.36 & 25.67 & 23.99 \\
Task Arithmetic & 16.42 & 17.67 & 22.61 & 14.15 & 30.89 & 25.48 & 11.03 & 5.47 & 20.29 & 14.47 & 36.10 & 49.15 & 26.88 & 22.28 & 44.35 & 40.14 & 26.07 & 23.60 \\
EMR-Merging & 16.34 & 17.90 & 20.97 & 13.02 & 29.12 & 24.81 & 11.70 & 5.80 & 20.47 & 14.53 & 35.65 & 49.08 & 26.26 & 21.25 & 43.28 & 39.11 & 25.47 & 23.19 \\
LoRaHub & 23.85 & 20.59 & 32.76 & 32.44 & 35.56 & 32.72 & 26.45 & 22.48 & 35.62 & 33.12 & 35.46 & 42.15 & 25.12 & 27.27 & 52.70 & 54.12 & 33.44 & 33.11 \\

\midrule 
Evo-Merging+ & \textbf{28.62} & \textbf{22.58} & \textbf{42.03} & \textbf{41.03} & \textbf{40.31} & \textbf{37.82} & \textbf{35.71} & \textbf{31.27} & \textbf{56.77} & \textbf{52.34} & \textbf{44.68} & \textbf{51.00} & \textbf{37.41} & \textbf{38.16} & \textbf{63.20} & \textbf{63.40} & \textbf{43.59} & \textbf{42.20} \\
\bottomrule 
\end{tabular}
\caption{Main Results for the Evaluation of In-Domain (ID) Robustness.}
\label{tab:method_performance_indomain_Robustness}
\end{table*}

\section{Hyperparameter Grid Search for Baseline Methods}
\label{sec:appendix_hyperparams}

This section provides a summary of the hyperparameter grid search conducted for the baseline methods across all datasets. 
This table \ref{tab:grid_search_summary} presents the optimal configurations and their corresponding best performance (Precision and F1-score in \%). The grid search for $\lambda$ was conducted over the set \{0.1, 0.3, 0.5, 0.7, 1.0\}, and for the density $p$ over \{0.5, 0.7, 0.9, 1.0\}. 

\section{More analysis about evaluation of Cross-task/domain Robustness }
\label{sec: More analyse about Evo-Merging on Muti-task merging}

To rigorously evaluate the robustness of our proposed merging algorithm, we introduce a set of five distractor tasks into the experimental setup. These tasks are intentionally selected from the Cross-Task benchmark, namely Relation Extraction (\texttt{RE\_NYT11}) and Event Extraction (\texttt{EE\_PHEE}, \texttt{EE\_CASIE}), to serve as sources of interference. Additionally, we incorporate two Entity Span Extraction tasks (\texttt{ES\_ACE\_2004}, \texttt{ES\_ACE\_2005}) to act as auxiliary tasks for Named Entity Recognition (NER). This setup is designed to stress-test the method's ability to handle parameter competition and mitigate negative transfer from cross-task learning.

The results presented in Table~\ref{tab:method_performance_indomain_Robustness} highlight a critical challenge in in-domain merging. Most baseline methods exhibit a significant performance degradation in the presence of the distractor tasks, a phenomenon akin to catastrophic forgetting. This decline can be largely attributed to the interference from heterogeneous task types, conflicting entity label sets, and the semantic divergence introduced by unrelated domains. In stark contrast, our Evo-Merging approach demonstrates remarkable resilience. By leveraging our specialized denoising and sign-flip scaling mechanisms, Evo-Merging effectively mitigates the negative influence from these unrelated tasks. It successfully filters parameter-level noise and harmonizes conflicting parameter directions, allowing it to distill shared, underlying knowledge from the diverse model pool. Consequently, rather than suffering from interference, our method thrives in scenarios with a greater number of models, achieving superior performance by leveraging the increased diversity.

\section{Further Analysis of Black-Box Merging}
\label{sec:more_analysis}

To further contextualize the advantages of our framework, we analyze its design principles against three distinct categories of methods within the challenging setting of Black-Box Merging (BMM) for 0ut-of-domain (OOD) tasks, particularly with massive model repositories.

\paragraph{Comparison with Task Vector-Based Merging Methods.}
Methods such as TIES-Merging, DARE, and Task Arithmetic have demonstrated considerable success in resolving parameter conflicts and combining model capabilities. They excel in in-domain, multi-task scenarios where the constituent models are pre-supposed to be relevant and beneficial. However, these methods exhibit a critical limitation in the BMM/OOD context: they lack an intrinsic mechanism to filter out irrelevant or detrimental models from a massive, noisy pool. Consequently, their application in such settings often necessitates laborious manual pre-selection of a small, promising subset of models and is highly sensitive to hyperparameter tuning. This reliance on pre-filtered, high-quality models leads to a catastrophic performance decline when confronted with a large-scale repository (e.g., 100+ models) containing task- or domain-irrelevant knowledge, a challenge our automated two-stage framework is explicitly designed to overcome.

\paragraph{Comparison with LoRaHub.}
LoRaHub represents a strong baseline for dynamic model composition via a weighted summation of LoRA modules. Nonetheless, as acknowledged in its own analysis and confirmed in our experiments, it also suffers from performance degradation when the model pool becomes excessively large. The primary reason is the absence of a dedicated, large-scale denoising mechanism. Its approach is insufficient to mitigate the compounded noise and signal conflicts inherent in a massive repository. In contrast, our Stage 1 (Sparsity-Based Denoising) directly confronts this issue by actively identifying and pruning noisy parameter spaces before fusion, enabling robust performance even as the model pool expands.

\paragraph{Comparison with Parameter-Efficient Fine-Tuning (PEFT) Methods.}
PEFT methods like LoRa and (IA)\textsuperscript{3} are highly effective for adapting a single model to new tasks with minimal parameter updates. However, their effectiveness diminishes in few-shot OOD generalization scenarios compared to merging approaches. This is because PEFT methods adapt from a single pre-trained source, whereas Evo-Merging is designed to harness the \textit{collective knowledge} distributed across a diverse repository of expert models. By integrating the rich "parametric memory" from numerous existing models, our framework activates superior generalization capabilities from only a few samples—a feat that is conceptually beyond the scope of standard PEFT.

\end{document}